\documentclass[10pt]{article}
\usepackage[a4paper,
            margin=1in]{geometry}
\usepackage{parskip}
\usepackage[activate={true,nocompatibility},final,kerning=true,spacing=true,factor=1100,stretch=10,shrink=10]{microtype}
\usepackage{setspace}
\usepackage{amsmath}
\usepackage{amssymb}
\usepackage{graphicx}
\usepackage{booktabs}
\usepackage{sectsty}
\usepackage{tabularx}
\usepackage[normalem]{ulem}
\usepackage{csquotes}
\usepackage[htt]{hyphenat}
\usepackage{nameref}
\usepackage{mathabx}
\usepackage{totcount}
\usepackage{epigraph}

\usepackage[hyphens]{url}
\usepackage{hyperref}
\hypersetup{
     colorlinks=true,
     linkcolor=black,
     filecolor=black,
     citecolor = black,      
     urlcolor=black}

\AtBeginEnvironment{displayquote}{\small \sffamily}

\usepackage{todonotes}

\usepackage[T1]{fontenc} 
\usepackage{mlmodern}
\usepackage[hang,flushmargin]{footmisc} 

\usepackage{etoolbox}
\AtBeginEnvironment{table}{\small}
\AtBeginEnvironment{tabular}{\small}
\AtBeginEnvironment{tabularx}{\small}
\AtBeginEnvironment{longtable}{\small}

\makeatletter
\renewcommand{\paragraph}{%
  \@startsection{paragraph}{4}{\z@}%
    {3.25ex \@plus1ex \@minus.2ex}%
    {-1em}%
    {\normalfont\fontsize{10.5}{12}\fontfamily{qhv}\selectfont\bfseries\protect\raisebox{-.225pt}}}
\makeatother

\makeatletter
\renewcommand{\maketitle}{
  \begin{flushleft}
    \fontfamily{qhv}\selectfont
    {\LARGE \textbf{\@title}} \\[2mm]
    \fontfamily{qhv}\selectfont \@author \\[2mm]
    \@date
  \end{flushleft}
}
\makeatother

\usepackage{caption}
\usepackage{subcaption}
\DeclareCaptionFormat{myformat}{#1#2#3\dotfill}
\title{Conflicting narratives and polarization on social media}
\author{\raggedright Armin Pournaki\textsuperscript{1,2,3}}
\date{%
    \small
    \textsuperscript{1}Max Planck Institute for Mathematics in the Sciences, Leipzig, Germany\\%
    \textsuperscript{2}Laboratoire Lattice, École Normale Supérieure - PSL - CNRS - Univ. Sorbonne Nouvelle, Montrouge, France\\%
    \textsuperscript{3}médialab, Sciences Po, Paris, France\\%
    [1ex]
     \today
}

\usepackage[style=authoryear-comp,
            natbib=true,
            backend=bibtex,
            giveninits=true,
            maxcitenames=2,
            maxbibnames=99,
            ]{biblatex}

\renewbibmacro*{cite:labeldate+extradate}{%
  \printtext[bibhyperref]{%
    \iffieldundef{note}
      {\printfield{year}\printfield{extradate}} 
      {\printfield{note}
       \addslash
       \printfield{year}\printfield{extradate}}}}       
\renewbibmacro*{finentry}{%
  \finentry
  \iffieldundef{note}
    {}
    {\newunit\newblock{ (Original work published \printfield{note})}}
  }

\AtEveryBibitem{%
    \clearfield{month} 
    \clearfield{day}   
    \clearfield{url}
    \clearfield{isbn}
    \clearfield{issn}
    \clearfield{note}
    \clearfield{urldate}
    \clearfield{address}
    \clearfield{editor}
    \clearlist{language} 
}

\AtBeginBibliography{%
  \renewbibmacro*{urldate}{} 
}

\DeclareFieldFormat{addendum}{\mkbibparens{#1}\nopunct}
\DeclareFieldFormat{edition}{\mkbibparens{#1}}
\DeclareFieldFormat[inbook]{title}{#1} 
\DeclareFieldFormat[article]{title}{#1} 
\DeclareFieldFormat[book]{title}{\mkbibemph{#1}} 
\DeclareFieldFormat[book]{volume}{(Vol. #1)} 
\DeclareFieldFormat{journaltitle}{\mkbibemph{#1}} 
\DeclareFieldFormat[inproceedings]{title}{#1} 
\DeclareFieldFormat[book]{edition}{(#1). } 
\DeclareFieldFormat{pages}{#1} 
\renewbibmacro*{title}{%
  \printfield{title}%
  \iffieldundef{edition}{}{\setunit{\addspace}\nopunct}%
}

\renewbibmacro{in:}{}
\renewbibmacro*{in:}{%
  \ifboolexpr{
    test {\ifentrytype{inproceedings}} 
  }
  {\bibstring{in}\intitlepunct}  
  {}            
}
\renewcommand*{\intitlepunct}{\nopunct}

\renewbibmacro*{volume+number+eid}{
  \addcomma
  \textit{\printfield{volume}}%
  \iffieldundef{number}
    {}
    {\mkbibparens{\printfield{number}}}%
  \setunit{\addcomma\space}%
  \printfield{eid}
}

\begin{document}
\newcommand{\node}[1]{\textbf{\small \textsf{#1}}}


\allsectionsfont{\usefont{T1}{qhv}{b}{n}\selectfont}

\maketitle

\vspace{7mm}
\noindent
{\fontfamily{qhv}\selectfont \large \textbf{Abstract}}\\[2mm]
{\small
Narratives are key interpretative devices by which humans make sense of political reality. In this work, we show how the analysis of conflicting narratives, i.e. conflicting interpretive lenses through which political reality is experienced and told, provides insight into the discursive mechanisms of polarization and issue alignment in the public sphere. Building upon previous work that has identified ideologically polarized issues in the German Twittersphere between 2021 and 2023, we analyze the discursive dimension of polarization by extracting textual signals of conflicting narratives from tweets of opposing opinion groups. Focusing on a selection of salient issues and events (the war in Ukraine, Covid, climate change), we show evidence for conflicting narratives along two dimensions: (i) different attributions of actantial roles to the same set of actants (e.g. diverging interpretations of the role of NATO in the war in Ukraine), and (ii) emplotment of different actants for the same event (e.g. Bill Gates in the right-leaning Covid narrative). Furthermore, we provide first evidence for patterns of narrative alignment, a discursive strategy that political actors employ to align opinions across issues. These findings demonstrate the use of narratives as an analytical lens into the discursive mechanisms of polarization.
}
\vspace{7mm}

\section{Introduction}
Narratives are key interpretive devices through which humans make sense of their environment \citep{Bruner1991,Polkinghorne1988}. They serve as lenses through which reality is experienced and told. In political science, narratives are described as meaning-making devices through which ``people make sense of their lives'' by ``construct[ing] disparate facts and weav[ing] them together cognitively to make sense of reality'' \citep{Patterson1998}. This sense-making step is often achieved by reducing the complexity of complex political events and relationships into ``easily understandable stories'' \citep{Groth2019}.
This reduction, which usually involves causal chaining of events and attributions of archetypical roles to political actors, such as protagonists or antagonists, allows the narrator to convey their political stance by selecting which events are included, which actors and relationships are highlighted and what role they play.
In political communication, this makes narratives powerful rhetorical devices that have been shown to affect beliefs and attitudes \citep{Green2000,Antinyan2024}.

Furthermore, narratives can serve as ways to define one's identity \citep{Somers1994}, and the identity of a group by delimiting the ``we'' from an outside ``they'' \citep{Hartley2015}. These identities, often defined through antagonization of an out-group, could be at the origin of larger political divides. Next to identities, \citet{Friedman2023} suggests that ideological polarization can be explained through the existence of diverging interpretive frames through which the world is experienced. We argue that narratives are a powerful analytical lens through which one can observe such diverging interpretive frames in text. This in turn provides insight into the central fault lines along which ideologically opposing camps divide. 

Only few studies have investigated the connection between polarization and narratives empirically. \citet{Jing2021} shed light on the role of  diverging narratives in the context of the Covid debate on Twitter in the US by computationally analyzing narratives of Democrats and Republicans. They find that (i) narratives diverge in focus, Republicans focusing on outside actors such as China, while the Democrats focus on health issues and financial support, and that (ii) both parties heavily antagonize each other in their narrative. Still within the topic of Covid, \citet{Zhao2023} show that the Twitter discourse on Covid in the US has become more polarized and aligned with party preferences over time. This points to another feature of narratives: the one of integrating disparate political events into pre-defined webs of meaning, and potentially instrumentalizing issues such as a pandemic to further fortify one's ideological views through strategic emplotment. Directly related is the question whether such strategies are actually successful in affecting opinions and beliefs. \citet{Antinyan2024} show in an experimental setting that selective exposure to different narratives about Covid may indeed affect participants' beliefs about the origins of the pandemic.

The present paper goes beyond previous approaches by systematically analyzing the conflicting narratives of opposing opinion groups across several salient political topics. Drawing upon previous work that empirically provides structural evidence for polarization and issue alignment in the German Twittersphere between 2021 and 2023 \citep{Pournaki2025}, we systematically extract and analyze textual signals of narratives \citep{Pournaki2024} of opposing ideological camps around the war in Ukraine, Covid and climate change in order to shed light on the following questions: What are the central discursive fault lines along which the left and right-leaning groups divide? What discursive strategies are employed in order to connect different issues such to facilitate the alignment of opinions? Approaching these questions using a combination of distant and close reading, this work surfaces the discursive points of tension between the two groups and reveal larger metanarrative patterns that could be at the origin of the observed issue alignment.

\section{Data}
This work uses a dataset of tweets previously introduced in \citet{Pournaki2025}. The dataset includes tweets on trending topics over the course of more than two years. More specifically, the authors collected tweets from 2021-03-29 to 2023-07-12 according to the following scheme: at the beginning of each day, a script was launched that collects the current ``trending topics'' (from now on referred to as ``trends'') in Germany using the Twitter Trend API (v1). By default, trends are personalized based on the account's Twitter/X usage. One can, however, disable the personalization by setting a specific location from which to draw the trending topics, which then yields ``popular topics among people in a specific geographic location'' \citep{X/Twitter2025}. The script was re-run every 15 minutes. At the end of each day, the authors counted the number of times each trending topic appeared during the day and kept the top 5 most frequent ones. This provided a proxy of the five most important trending topics for that day. The Twitter Search API (v1) was used to collect German-speaking tweets using the exact trend keyword as a query on the day it trended and the day after (48hrs). Tweets were collected using a single Twitter API key, collecting tweets for maximally 24 hours every day. Each collected trend is one sub-dataset. This results in 3007 trending topics collected. Trends were merged if the exact same phrase occurs within a time window of 1 day. After merging, there are 2693 sub-datasets. These contain a total of 19,105,532 tweets.

\section{Methods}
\begin{figure}[t]
  \centering
  \includegraphics[width=\textwidth]{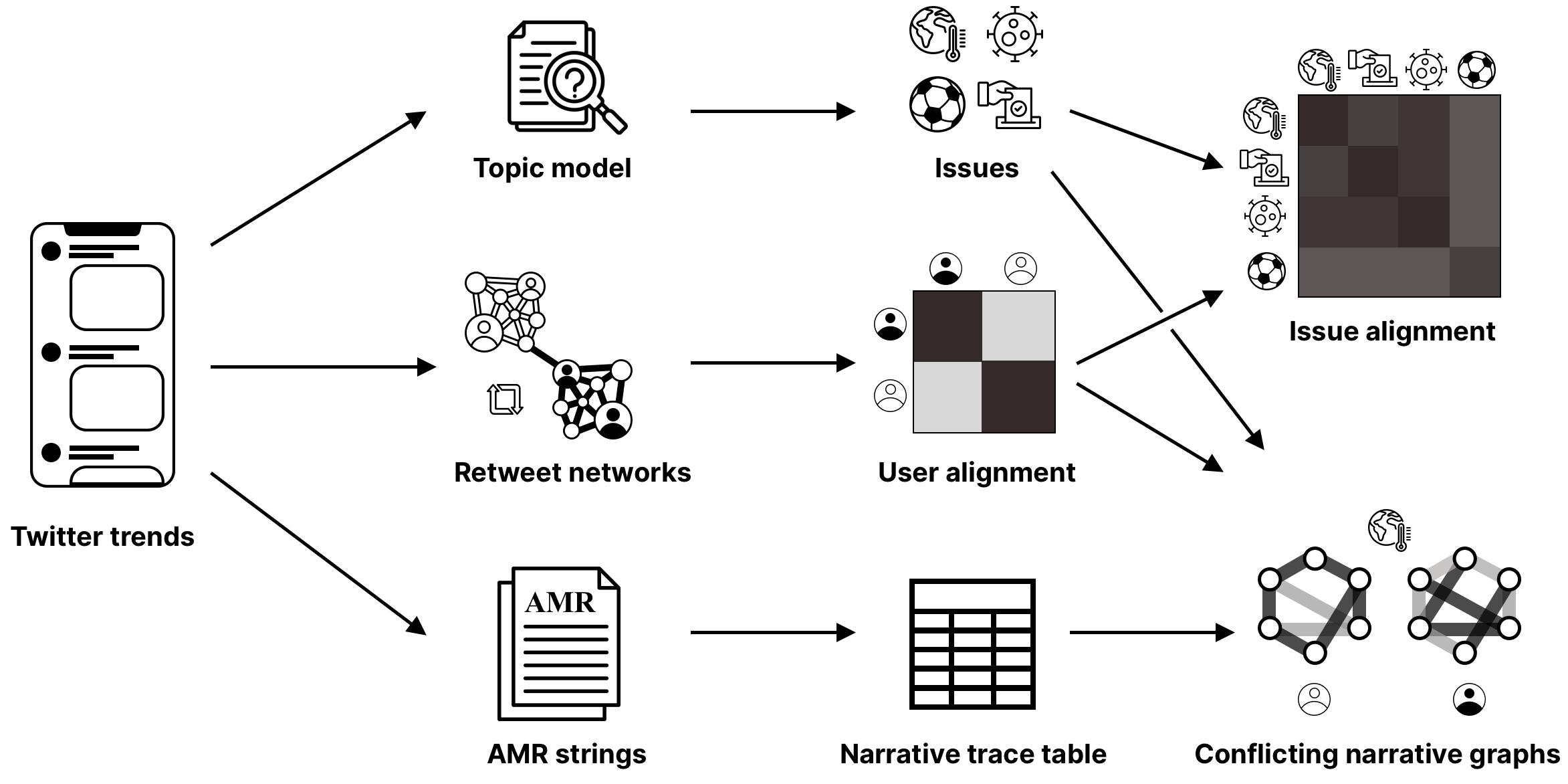}  
  \caption[Twitter Trends Analysis Pipeline]{\textbf{Analysis pipeline.} The raw text data of tweets is processed in a topic model to extract the main issues discussed, which is then used to assign an overall issue to each trend. In parallel, retweet networks are constructed and clustered to extract the opinions of users for each trend. From this set of partitions, the most prominent users are extracted (influencers and multipliers) and a \textit{user alignment} matrix is computed that quantifies how systematically pairs of users appear in the same cluster across the dataset. Importing the issue labels from the topic model analysis allows to compute the user alignment separately for each issue, which is then used to generate the issue alignment matrix that measures how similarly two issues sort these users into opinion groups. The results from the structural analysis of retweet networks allows to divide the textual corpus into ideological clusters. After processing the tweets using the narrative pipeline described in \citet{Pournaki2024}, we systematically compare the actantial networks to uncover conflicting narratives for a selected range of salient topics (War in Ukraine, Covid and climate change).}
  \label{fig:pipeline}  
\end{figure}
\subsection{Structural analysis of polarization}
The structural analysis of polarization on Twitter was previously presented in \citet{Pournaki2025} and is described in the upper two rows of Fig.~\ref{fig:pipeline}. We summarize it very briefly here in order to contextualize the present findings.

\subsubsection{Extracting issues from tweets}
First, all tweets were processed through a BERTopic-based pipeline \citep{Grootendorst2022} which allows to inductively surface the latent topics present in the text by clustering tweets in a semantic space given by pre-trained sentence embedding models. The assumption is that clusters in that space correspond to topic clusters. These topics were then manually grouped into larger issues, such as Covid (including topics such as vaccination, masks, school closures), or climate change (including topics such as activism, meat consumption, ...). Each trend could then be assigned one of those larger issues, based on the majority of tweets it contains. Table~\ref{tab:topicmodel} shows the issues extracted through this method and their prevalence across the corpus.

\subsubsection{Extracting opinions from tweets}
In parallel, a retweet network was extracted from each trend, where nodes are Twitter accounts and a directed link is drawn from account $i$ to $j$ if $i$ retweets $j$. Since retweets are considered endorsement, densely connected clusters in such networks correspond to opinion clusters in the underlying political debate \citep{Conover2011a,Gaisbauer2021,Gaumont2018}. Conceptually, a retweet network with two clusters can therefore be treated as an opinion poll. A user can take either the stance $-1$ or $1$ on a given issue spanned by a trending topic. An SBM-based approach \citep{Peixoto2019} provides the cluster assignment for each node in each network. This allows to measure, across the dataset, how often a given pair of users are in the same opinion cluster. Computing a measure of user alignment and clustering the resulting alignment matrix shows that the user base is largely divided into two camps: one left-leaning and one right-leaning, which were identified by manually examining the most active users in each camp. 

The global clustering into left/right is used in order to assign each tweet to one of the two camps, based on the majority of retweets it recieved. This allows to generate, for each issue, two sub-corpora: one of left-leaning and one of right-leaning tweets, that are analyzed separately for conflicting narrative signals.

\subsection{Extracting conflicting narratives}
\label{subsec:extracting_conflicting_narratives}
In order to analyze conflicting narratives of various topics in Twitter discussions, this work builds upon the extraction of \textit{narrative signals} developed in \citet{Pournaki2024}. The central idea is that narratives are co-created and often not explicitly stated in full in the text. Therefore, the challenge of extracting narratives from digital traces is to identify key narrative signals, in the form of actors, their relationships, and their goals/motives. These signals are then reassembled into \textit{actantial networks}, where nodes are actors, and links denote the nature of their relationship as conflictive, supportive, or neutral. These networks are typically very large and contain information that is not necessary for the analysis of political narratives. To get insight into the most prominent narrative signals, the nodes in the networks are therefore filtered using centrality measures (in-/out-degree, betweenness). Finally, we recall that each edge in the network arises from a collection of textual passages in the original data\footnote{The sentence ``Russia attacked Ukraine.'' would, for instance, contribute to a conflictive link from \node{\footnotesize Russia} to \node{\footnotesize Ukraine}.}. Pointing precisely to these passages allows to contextualize each of those actantial links and add a hermeneutic layer to the computational analysis. The approach therefore combines a fully inductive, machine-learning based extraction of narrative signals from raw text with an interpretive layer of close reading, which allows to reconstruct the underlying (meta-)narratives. 

In practice, the approach consists in first transforming text into a structured representation using the semantic parsing framework of Abstract Meaning Representation (AMR) \citep{Banarescu2013}. This framework represents the meaning of sentences as directed acyclical graphs, where the root is the verb of the sentence, and the leaves are given by the arguments (e.g. subject/object). The corpus is therefore parsed into a set of AMR graphs, which are then transformed into a tabular representation, which can easily be queried for narrative signals, and from which actantial networks can be generated (we refer to the reader to \citet{Pournaki2024} for a detailed description of the processing steps involved). 
Since the AMR parser in its current state does not handle German, we by translate our full corpus of original tweets into English using the transformer-based model \texttt{mbart-large-50-many-to-many-mmt}\footnote{\url{https://huggingface.co/facebook/mbart-large-50-many-to-many-mmt}}, that provides a good tradeoff between efficiency and accuracy \citep{Tang2020}. We then parsed the left and right-leaning subcorpora to analyze the systematic differences in narratives of the opposite camps across topics, with respect to involved actors, their ascribed actions, as well as relationships in the form of actantial networks. A link from node $i$ to $j$ in the actantial network has a weight $w(i,j)$ which denotes the number of times a relationship from $i$ to $j$ has been (re)tweeted in the corpus, and a score $\sigma(i,j) \in [-1,1]$ which denotes how supportive ($\approx+1$), conflictive ($\approx -1$) or neutral ($\approx 0$) the relationship is. This score is computed based on all the individual interactions from $i$ to $j$, which, in \citep{Pournaki2024} are classified using verb category dictionaries \citep{DiFabio2019}.

In this work, we extend the narrative signal extraction to allow for the systematic extraction of conflicting narratives, and to better fit the Twitter data at hand, by adding the following two processing steps:

\subsubsection{LLM-assisted actantial link labeling}
In the actantial network, each node is an actor and a directed, weighted, scored link is drawn from actor $i$ to $j$ where the weight is defined as the
sum of retweets of all tweets in which the edge is contained, and the score denotes whether the link is supportive (positive values), conflictive (negative values) or neutral (zero). In \citet{Pournaki2024}, the approach to label a given relationship is based on a pre-defined categorization of verb families using the VerbAtlas dictionary \citep{DiFabio2019}, where each verb family was assigned one of the three categories\footnote{The Verbatlas family {\sc attack\_bomb}, which contains verbs like ``invade'' would, for example, be labeled as conflictive.}. 
This bears a number of limitations which mainly revolve around the question of \textit{context}, which potentially plays an important role in the determination of the conflictive or supportive nature of the relationship. In the present setting, the context consists of the actors involved, and the preceding and following sentences in the tweet.

We identify Large Language Models (LLMs) as a potential tool to include this broader context into the analysis. Previous work has shown their potential in text annotation tasks, and in particular that their accuracy can exceed the one of human annotators \citep{Gilardi2023,Aldeen2023,Alizadeh2024,Ziems2024,Bojic2025}, even though this depends on the domain \citep{Lu2023,Nasution2024}. Aside from providing the full tweet as context to the LLM, this approach allows to use the original, German-speaking tweet for the inference of the relation type. This can mitigate potential mistakes in the translation that could have occurred at an earlier step of the pipeline.

We use the open weight model \texttt{Phi-4} \citep{Abdin2024}, forcing the model to produce JSON output using structured/guided generation \citep{Willard2023}, which highly facilitates further processing and makes the pipeline more efficient by decreasing inference time. We report the prompt in the Appendix \ref{sec:prompt} and validate and assess the benefits and limitations of the LLM-based approach in the Appendix \ref{sec:llm_validation}.

\subsubsection{Systematic extraction of conflicting narratives}
We approach the extraction of conflicting narratives by systematically comparing links in actantial networks. Let $G_{l}(V_{l},E_{l})$ be the actantial network extracted from left-leaning tweets and $G_{r}(V_{r},E_{r})$ the actantial network extracted from right-leaning tweets. Take two edges $e_1 = (i,j,w_l(i,j),\sigma_l(i,j)) \in E_l$ and $e_2 = (i,j,w_r(i,j),\sigma_r(i,j)) \in E_r$, where $w$ denotes the weight and $\sigma \in [-1,1]$ denotes the edge score. We assume conflicts between two narratives to surface when relationships between actors are narrated in an opposite way (i.e. supportive in one narrative vs. conflictive in the other). Therefore, if $\text{sign}(\alpha_l(i,j)) \neq \text{sign}(\alpha_r(i,j))$, we assume the underlying edge to correspond to a narrative trace that points to a fault line in the underlying debate. Finally, we examine the actantial networks made of only those links that point to potential conflicts. Paired with a close reading of relevant tweets associated to those links, the approach allows to reconstruct the diverging lenses through which the two camps view political reality.

\section{Results}
\label{sec:results}
We focus on a subset of prevalent issues during the data collection period, namely the war in Ukraine, Covid and climate change. The goal of this analysis is twofold: (i) discover the underlying conflicting narratives that might explain the observed polarization, and (ii) discover the discursive mechanisms by which these different issues may be connected. Using the methodology described in \ref{subsec:extracting_conflicting_narratives}, we start by analyzing each of the three issues separately, and then move to a combined analysis that reveal first evidence of larger meta-narrative patterns.

We start by analyzing each of the three issues separately. We generate the subset of tweets of each camp and extract the actantial networks. We start by apprehending \textit{identity narratives} as out-going ego networks of the node \node{we} for both camps. Then, we extract \textit{conflict networks}, subsets of the actantial networks that contain the most central actants (selected by ranking the nodes in the full actantial networks by in-degree/out-degree and betweenness centrality and keeping the relevant ones within the top 100), and that solely contain links that have an opposing score between the two camps. These conflict networks surface the central fault lines in the polarized discussions.

These networks can be concieved as re-assembled narrative signals that, by themselves, provide a distant overview of the narrative content in the underlying textual corpus, and surface the discursive points of tension between the communities. To fully apprehend the underlying narratives, it is necessary to complement the network representations with a close reading approach of a subset of relevant tweets that led to the extracted links. In the following section, we demonstrate this guided close reading protocol: first, we extract the different types of narrative networks (ego networks of the nodes \node{we} and networks of most important actors for both camps). Then, we contextualize the links by reading and interpreting the 5 most retweeted tweets of each link (see Appendix for a selected list of close-read tweets). Finally, we summarize the conflicting narratives for each of the three investigated issues, and provide first insights into the common patterns observed across them that may explain the observed issue alignment.

\subsection{War in Ukraine}
One central topic in the timerange we observed was the Russian invasion of Ukraine. Russia began increasing its military presence near the border from March/April 2021, before launching the invasive attack on Ukraine's territory on February 24 2022. This has since led to the deadliest war in Europe since World War II, causing large numbers of military and civilian casualties and displacements. The war is also fought on the level of (mis)information and narrative, which this section sheds light on by systematically analyzing the narrative signals of left- and right-leaning Twitter communities.

\subsubsection{Identity narrative}
\begin{figure}[t]
  \includegraphics[width=\textwidth]{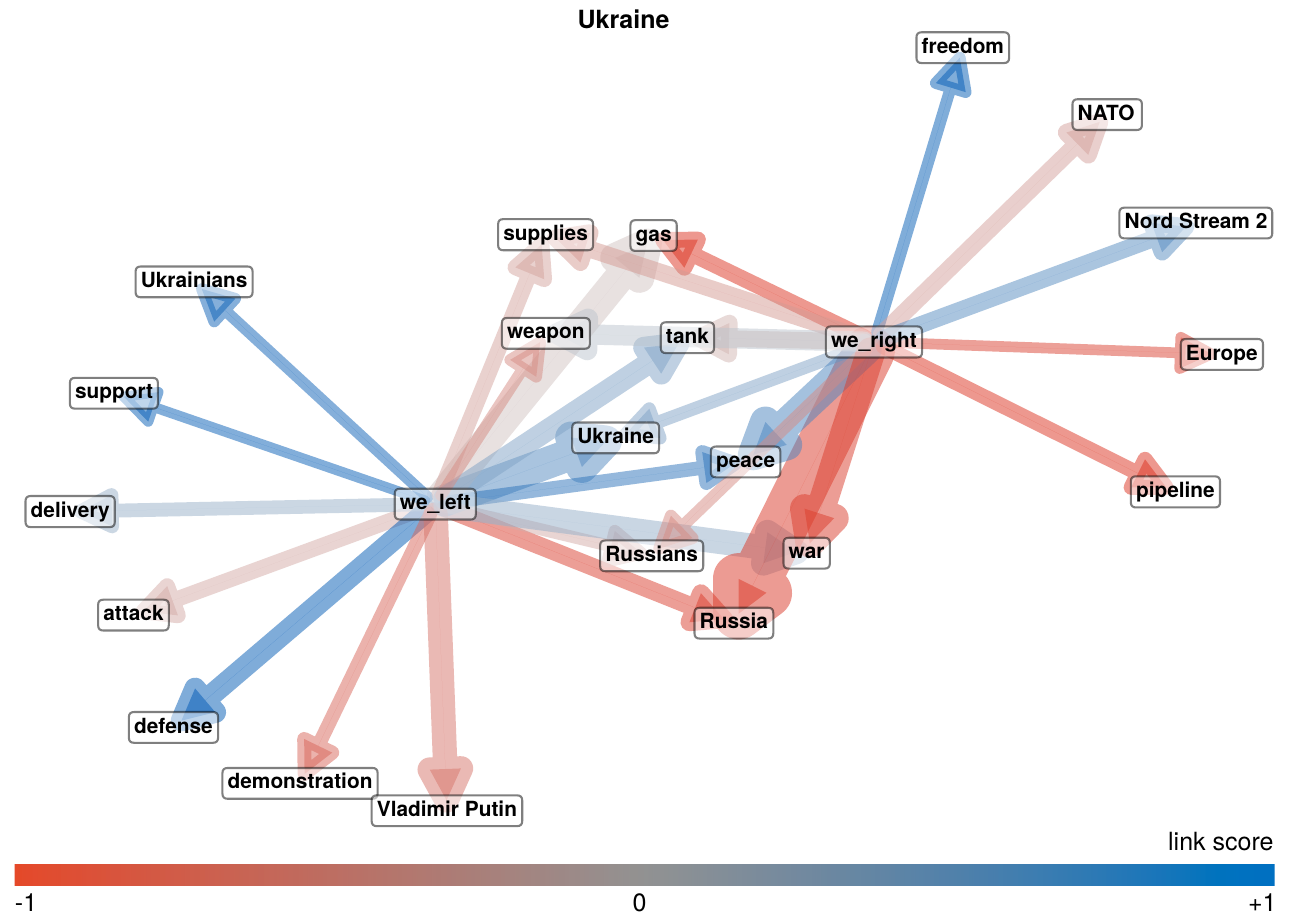}
  \caption[Identity narrative network (Ukraine)]{Identity narrative network extracted from trends related to Russia's aggression of Ukraine. Link width is proportional to weight, link color reflects the link score (red$=$conflictive, blue$=$supportive, grey$=$neutral, color bar only plotted here for subsequent actantial networks). Links reflect relationships retweeted at least 500 times.}
  \label{fig:wegraph_ukraine}
\end{figure}
Figure \ref{fig:wegraph_ukraine} shows the merged ego-networks of the node \node{we} for the topic Ukraine. The links are filtered by the sum of the number of (re)tweets that contain the relationship: each edge needs to be retweeted at least 500 times. We see that certain nodes are only connected to either one of the \node{we} entities, while certain nodes are connected to both. The former represent actants emplotted in relation to only one of the groups, while the latter represent actants to which relationships can be contested across the two groups.

Examining the network and close reading the most retweeted tweets associated to each edge allows us to reconstruct the narratives built around each of the collective identities (see Appendix \ref{sec:appendix:closereading} for the corresponding tweets). The left-leaning camp identifies itself through strong solidarity and support for Ukraine (supportive links towards \node{Ukrainians}, \node{support}, \node{defense}), which implies delivering \node{weapons} to Ukraine. \node{Vladimir Putin} is strongly condemned, the corresponding tweets show that he is explicitly blamed for starting the war. Furthermore, we observe a conflictive stance towards \node{demonstrations}, which refer to ``demonstrations for peace'' organized by groups with pro-Russian tendencies in Germany.

Turning to the nodes connected to the right-leaning \node{we}, we observe supportive relationships to \node{freedom} and \node{Nord Stream 2}. Freedom is often evoked in right-wing narratives, and in the context of the Ukraine war it is connected to freedom of action (not to enter the war with Russia), and freedom of expression (for non-interventionist voices promoting a pro-Russian narrative). We observe conflictive relationships to \node{NATO} and \node{Europe}, which are considered culprits in pushing Russia into an unavoidable defense war (described more explicitly in the following subsection).

Looking at the contested nodes between the two camps, we observe that both parties exhibit a supportive relationship to \node{peace}. The tweets connected to the link reveal different conceptions as to how to reach peace. On the left, peace can only be reached by full support for Ukraine, while the right promotes the possibility of peace through non-intervention and reiterated phrases like ``Weapons cannot bring peace.''. Looking at the connections to \node{war}, the left-leaning narrative exposes the shortcomings and lacking sanctions and deals that allow to ``finance Putin's war'' (hence the non-negative link). The right-leaning camp on the other hand strictly opposes Germany's involvement in the war. 

\subsubsection{Conflicting narratives}
\begin{figure}[t]
  \includegraphics[width=\textwidth]{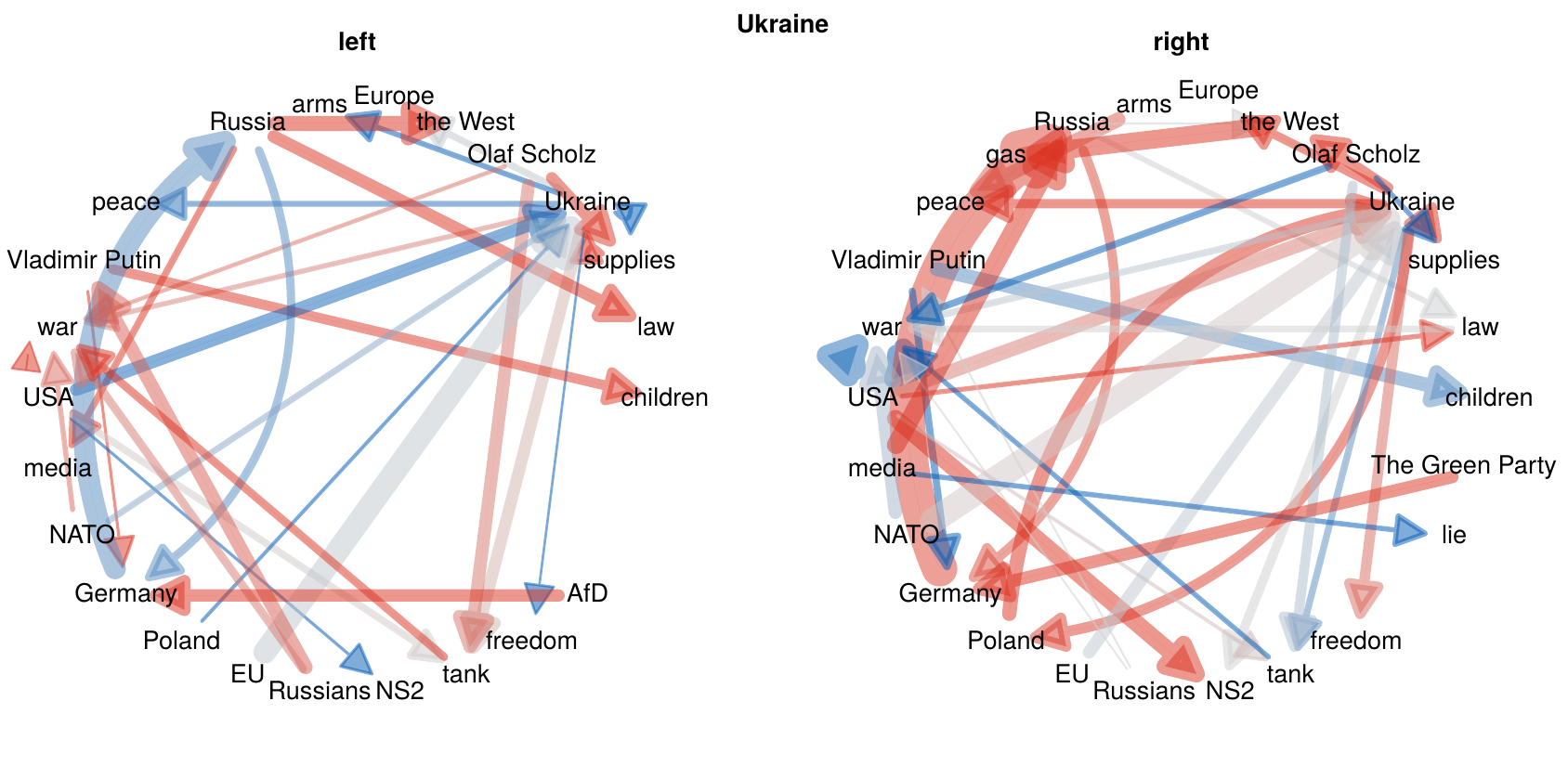}
  \caption[Conflict networks (Ukraine)]{Conflict networks extracted from trends related to Russia's aggression of Ukraine. Each link has opposite sign in the left/right network. 
  }
  \label{fig:conflictgraph_ukraine}
\end{figure}
Figure~\ref{fig:conflictgraph_ukraine} shows the actantial network of central nodes in both left and right-leaning camps. Only links with different sign of scores for left/right are drawn.
This highlights the fundamental interpretive differences of political events between the two camps. While, on the left, the \node{USA} and \node{NATO} are trying to help \node{Ukraine} who is being attacked by \node{Russia}, the right-leaning camp identifies them as the main culprits in this war. This difference leads to other rifts, such as the divisive stance towards weapon supplies insinuated by the difference in connecting weapons and peace: in the right-leaning cluster, weapons can never lead to peace, whereas the left-leaning cluster has a more nuanced stance and supports weapon supplies to protect Ukraine's sovereignty.

Furthermore, we observe that both sides blame the ideological opponents, such as the \node{media} and the \node{Green Party} by the right, and the \node{AfD} by the left. In particular, the \node{media} is accused of lying and promoting anti-Russian propaganda in the right-leaning camp, while the left accuses \node{Russia} of silencing independent media. The role of \node{Russia} as a lawful state is contested as well: while the left-leaning camp emphasizes that the attack is against international \node{law}, the right-leaning narrative suggests that Russia is merely (lawfully) defending itself against the threat of \node{Ukraine}, \node{USA} and \node{NATO}.

We observe another example of conflicting interpretations of reality when we turn our attention to the link from \node{Vladimir Putin} to \node{children}. In March 2023, the International Criminal Court arrested a warrant against Putin, claiming he is ``responsible for the war crime of unlawful deportation of population (children) and that of unlawful transfer of population (children) from occupied areas of Ukraine to the Russian Federation'' \citep{ICC2023}. This allegation is echoed in the left-leaning camp, while the right-leaning camp claims that Putin \textit{saved} those children from war zones, thus portraying him as a hero rather than a criminal. This example shows how the same event can serve to narrate an actant's position in opposite ways, depending on which position fits the larger ideological narrative.

\subsection{Covid}
We now turn our attention to another central trending topic in the dataset: the Covid pandemic. The first trending hashtag in the dataset appears on 2021-03-30 (``\#harterLockdownJetzt'') and the last on 2023-06-17 (``Lauterbach''). Across the timerange, we find discussions around the various facets of the pandemic, from counter-measures like lockdowns, schools closures, home office to prevention like masks, testing and vaccination. These sparked polarized debates about individual freedom, trust in science and the government.

\begin{figure}[t]
  \includegraphics[width=\textwidth]{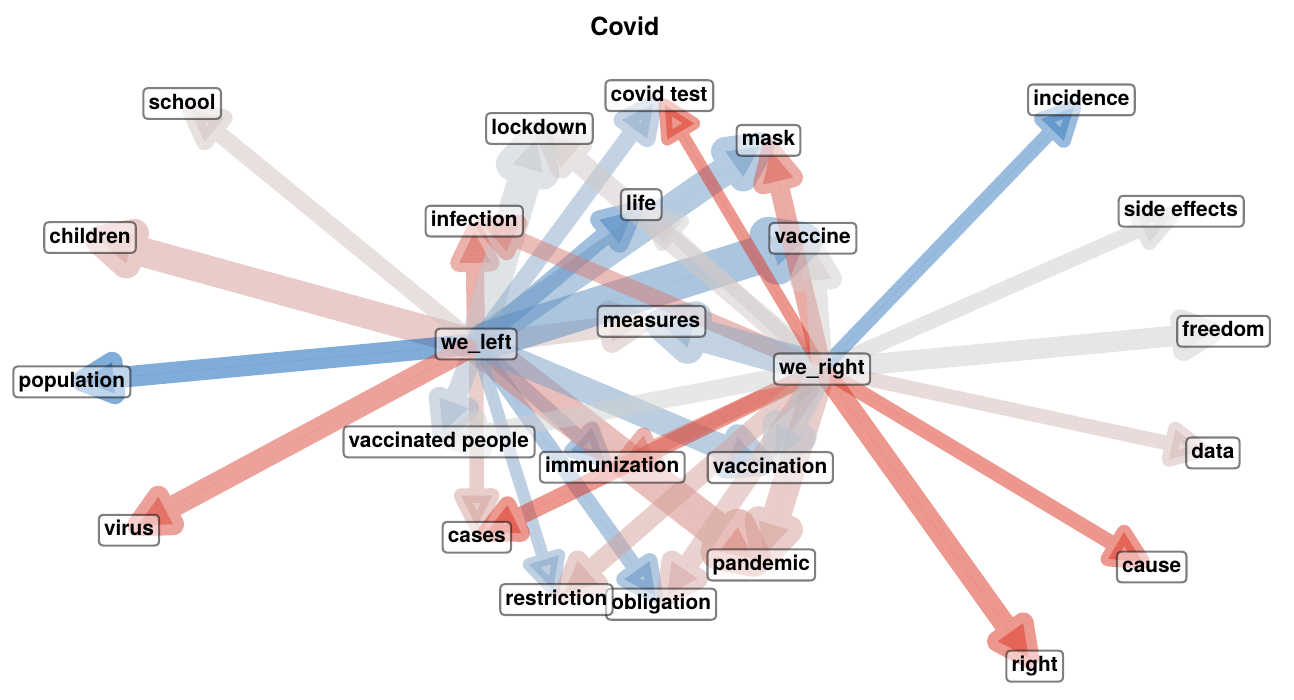}
    \caption[Identity narrative network (Covid)]{Identity narrative network extracted from trends related to Covid. Link weight threshold 500.}
  \label{fig:wegraph_covid}
\end{figure}
\subsubsection{Identity narrative}
Figure~\ref{fig:wegraph_covid} shows the identity narrative network around the Covid pandemic. We observe that the left-leaning cluster identifies itself through the positive stance towards measures to prevent virus spread. This includes \node{masks}, \node{tests}, \node{lockdowns}, \node{restrictions}, and \node{vaccinations}. The right-leaning camp strongly opposes these measures while referring to fundamental human \node{rights} and misguided, politicized science.
There is a focus, in the left-leaning narrative, on solidarity and compliance to fight the disease together, while the right-leaning narrative focuses on individual \node{freedom}. These differences in narrative turn masks or vaccines into markers of ideological identity: wearing masks, getting vaccinated can act as signals of ideological leaning.
\subsubsection{Conflicting narratives}
\begin{figure}[t]
  \includegraphics[width=\textwidth]{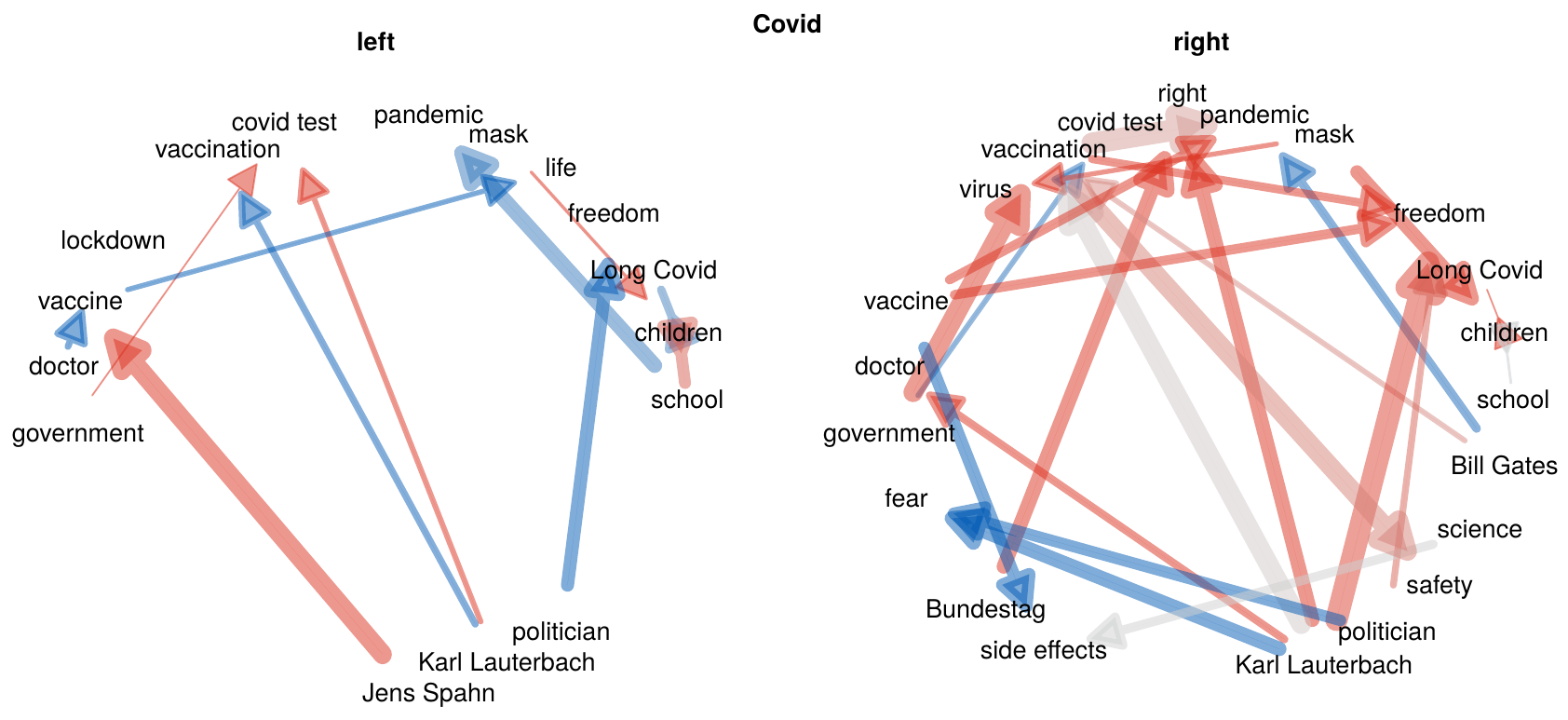}
  \caption[Conflict networks (Covid)]{Conflict networks extracted from trends related to Covid.
  }
  \label{fig:conflictgraph_covid}
\end{figure}
Figure~\ref{fig:conflictgraph_covid} shows the conflict networks for the Covid pandemic. We observe divergent views on the effectiveness of the \node{vaccination} and its \node{side-effects}, and trust in \node{science}. The vaccine's safety is seriously doubted in the right-leaning narrative, while it is considered a necessary solution to fight the pandemic on the left. We observe the interpretation of \node{masks} and other measures as harmful attacks on fundamental \node{rights}. Different political actors are antagonized: on the left, \node{Jens Spahn} (minister of health from 2018 to 2021) for his allegedly poor management of vaccine distribution, and on the right, \node{Karl Lauterbach} (minister of health 2021 to 2025) for his strong pro-vaccine stance and his alleged attacks on \node{freedom} and fundamental \node{rights}.

Looking at \node{children}, we observe differences in what each camp considers harmful for them: the left-leaning camp warns of ``sacrificing'' children by keeping \node{schools} open and increasing their chances of being infected, while the right-leaning camp warns of the potential psychological harms connected to \node{masks} and isolation.

We also observe certain actors that only play a role in the right-leaning narrative: for one, \node{fear} is being mongered by politicians in order to get the population to comply with the harsh measures. Then, \node{science} proves the importance of the vaccine's \node{side-effects}, which are not discussed in the left-leaning camp. Finally, we find traces of the conspiracy theory around \node{Bill Gates}' involvement in the orchestration of the pandemic for personal profit, which does not exist in the left-leaning narrative. 

In the right-leaning camp, we observe traces of a recurring general narrative about individual freedom: in this case, it is infringed upon by the political establishment through harsh measures. This theme also connects Covid to the topic of climate change, as will become visible in the following section.

\subsection{Climate change}
The climate change topic is a multi-faceted issue in the dataset and spans trends from the beginning of the data collection on 2021-03-29 to 2023-07-12. The related sub-topics are on one hand weather events such as the heavy floods in July 2021 (\#Hochwasserkatastrophe, 2021-07-18), discussions about speed limits on highways (Tempolimit, 2021-10-18), climate activism (\#LetzteGeneration, 2023-05-26), the climate summits (\#COP27, 2022-11-10), and the town of Lützerath that was destroyed to build a coal mine and sparked large-scale demonstrations (\#Luetzerath, 2023-01-14). The subsequent analysis shows that there are fundamental differences between the two camps as to how extreme weather events, as well as climate change and climate activism, are interpreted and emplotted.
\subsubsection{Identity narrative}
\begin{figure}[t]
  \includegraphics[width=\textwidth]{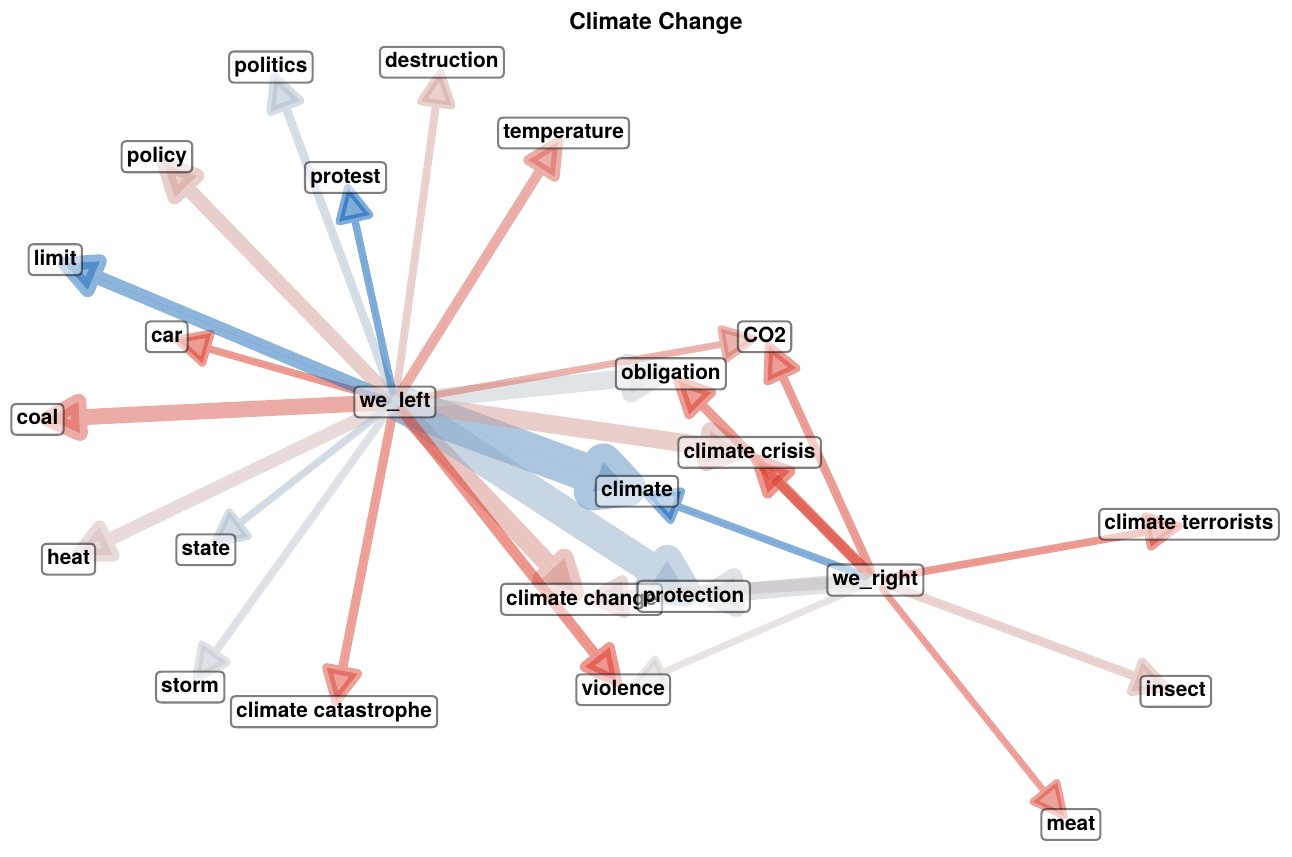}
  \caption[Identity narrative network (climate change)]{Identity narrative network extracted from trends related to climate change. Link weight threshold 250.}  
  \label{fig:wegraph_climatechange}
\end{figure}
Figure~\ref{fig:wegraph_climatechange} shows the identity narrative networks of climate change. One main difference between the left-leaning and the right-leaning camp arises from the interpretation of \node{climate change} as an actual threat. This shows itself through the use of the word \node{climate catastrophe}. From this standpoint, the left-leaning cluster derives a negative stance towards \node{coal}, a positive stance towards speed and CO\textsubscript{2} emission \node{limits}, and climate \node{protests}. On the right, the role of CO\textsubscript{2} is relativized, and the emphasis is laid on the illegal activities of \node{climate terrorists}. Furthermore, there is a focus on the consumption of \node{meat}, induced by the fear that the supermarket chain Lidl announced an increase in plant-based products (the negative link arises from sentences like ``We will have to renounce meat''). Along similar lines, we find traces of a conspiracy narrative around political elites forcing the regular population to eat \node{insects} while they themselves eat steaks on their yachts. This echoes previously identified narratives around individual freedom that is being infringed upon by political elites.

\subsubsection{Conflicting narratives}
\begin{figure}[t]
  \includegraphics[width=\textwidth]{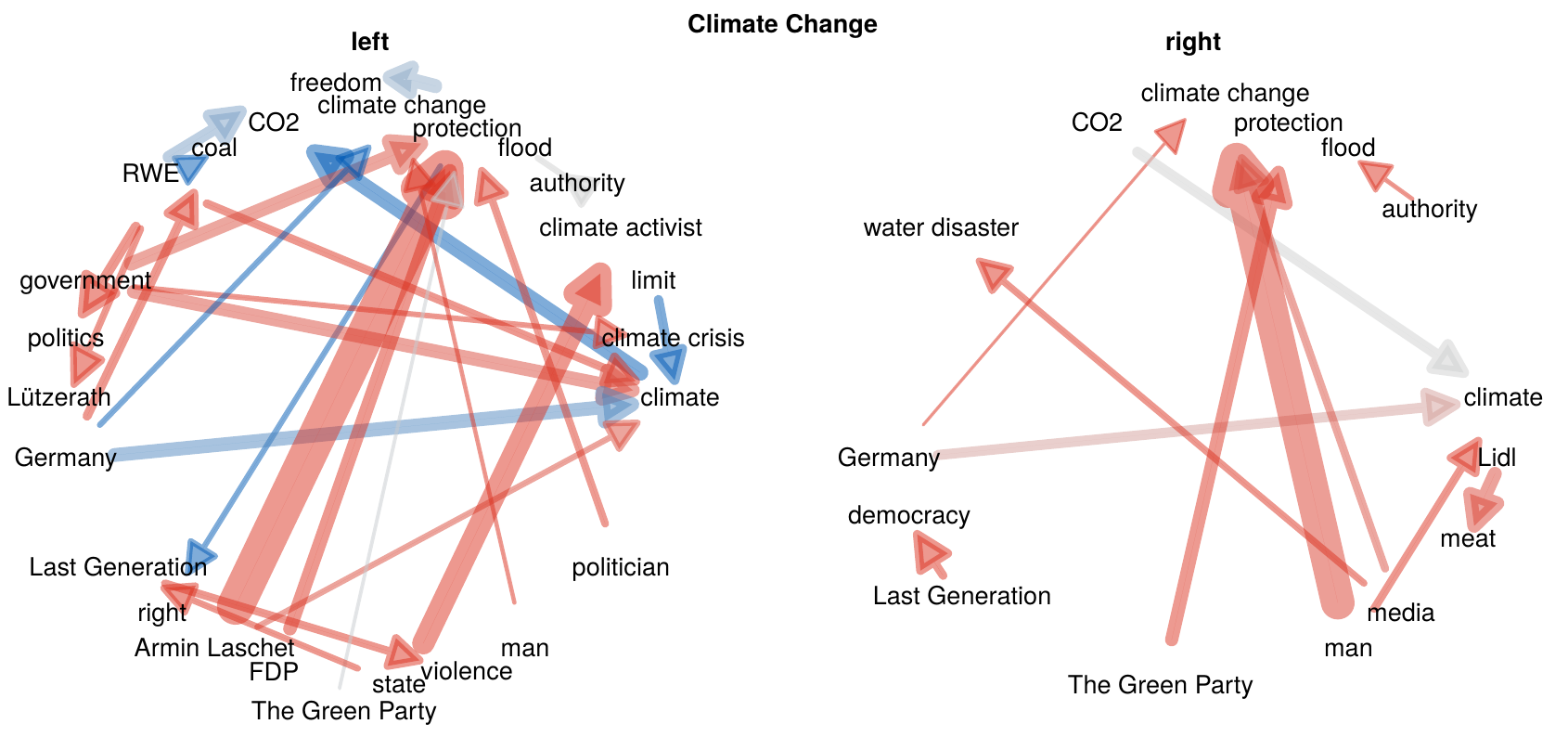}
  \caption[Conflict networks (climate change)]{Conflict networks extracted from trends related to climate change.
  }
  \label{fig:conflictgraph_climatechange}
\end{figure}
Figure~\ref{fig:conflictgraph_climatechange} shows the conflict networks of climate change. We observe a central difference between the two camps with respect to the impact of climate change on society. While the left considers \node{floods} a direct consequence of \node{climate change}, the right does not acknowledge this connection. Similarly, we observe differences in the evaluation of the dangers of \node{coal}.

Staying within the right-leaning narrative, we observe that the \node{Green Party} and the fear-mongering \node{media} are antagonized, as in the war in Ukraine. \node{Lidl} is evoked in the right-wing narrative due to allegations that the supermarket chain will reduce sales of \node{meat} products.
Furthermore, one central interpretive conflict lies in the role \node{Germany} can play in the larger scheme of \node{climate change}: for the right, Germany cannot respond to stop climate change and should therefore focus on other issues, while the left believes in Germany's responsibility to tackle the issue. This may lead to conflicting conceptions about the \node{Last Generation} and \node{climate activists}. On the right, their actions are heavily condemned as illegal and anti-democractic, while the left supports their efforts. Finally, we note that climate change is not a pressing issue for the right, while the left is convinced that it is one of \textit{the} central issues that need to be tackled.

The analysis of conflicting narratives for each of the three issues revealed the central fault lines along which the debates are polarized. While these fault lines are usually topic-specific, such as the diverging interpretations of NATO's role in the war in Ukraine, the guided close reading analysis surfaced some elements common to all three issues, such as antagonizations of the media in the right-leaning cluster. The next section shall provide a first approach to analyze the narrative elements that are used to bind issues together and that may be at the origin of the observed issue alignment.

\subsection{Issue alignment}
We use the insights gained from the analysis of individual issues to generate first hypotheses about the role of narratives for issue alignment.
One way in which observe issues to be aligned are through the invokation of recurring actors as protagonists or antagonists. Focusing on the connection between climate change and Covid, we find that the left-leaning camp systematically holds up their trust in \node{science}:

\begin{displayquote}
``Why does politics not listen to science?? \#Laschet \#Hochwater disaster \#Climate disaster \#Climate crisis'' (2021-07-16 09, 79 retweets)\\[2mm]  
``I’ve been vaccinated 4 times, I’ve vaccinated over 3,000 patients in my clinic, I’ve set up a fever ambulance and I’m doing social media education. \#DasHabeIchGetan I want politics to listen to science and provide me with vaccines. I don’t play golf. « NatalieGrams: Uncle brings me an idea. So write down under \#DasHabeIchGetan what you’ve done to slow down the pandemic. Maybe politics will finally understand what it has to do! — »'' (2021-11-28 18, 136 retweets)\\[2mm]
``“Politics must finally learn to face science on an eye-to-eye basis.” — Christian Althaus, Bernese epidemiologist, after leaving the Corona Taskforce of the Federal Government.'' (2021-04-05 08, 74 retweets)
\end{displayquote}

On the right, we observe that (mainstream) \node{science} is discredited:

\begin{displayquote}
  ``What if politics and science are on the same side of climate change as Corona?'' (2023-05-20, 621 retweets)
\end{displayquote}

We also observe signals of recurring narratives around \node{fear} and \node{freedom} that connect issues together. This leads to analogies between the government's approach to Covid and climate change policy, as exemplified through the emerging keyphrase ``climate lockdown'', a powerful rhetorical tool that binds the issues together:

\begin{displayquote}
  ``Please tell me that the people are not so submissive again and would accept a climate lockdown?!'' (2023-06-16 07, 215 retweets)\\[2mm]
``Are you already looking forward to the CO2 lockdown and the QR code on your phone, without which you will never be able to go anywhere? \#Lanz \#RicardaLang @ZDF \#GreenGarbage'' (2023-05-18, 555 retweets)  
\end{displayquote}

On both camps, we observe the antagonization of ideologically opposite political parties. On the left, the \node{AfD} is constantly evoked as a threat to Germany, while on the right, the \node{Green Party} (often mentioned as the ``prohibition party'', highlighting their politics as a threat to individual freedom) plays that role as the central antagonist. This is exemplified through the villanization of the Green politician \node{Karl Lauterbach} as a central antagonist in several topics:

\begin{displayquote}
``The stupid Karl \#Lauterbach announces a heat protection plan, although much more people die in cold than in heat. Therefore, a cold protection plan would make more sense: \#Open Nordstream2 and \#Restart nuclear power plants.'' (2023-06-13 23, 151 retweets)
\end{displayquote}

Furthermore, we observe the general distrust in \node{media} in both topics:

\begin{displayquote}
  ``Fact: During \#Corona, we were deceived, framed by the media, trumpeted by purchased experts... Other opinions (which were the right ones) were suppressed, made impossible, shamed... EXACTLY THE SAME takes place again on the topic of \#Climate Change! \#Berlin2030'' (2023-03-25 18, 335 retweets)\\[2mm]
``Here speaks a courageous nurse from New Zealand. `No one listens to us and the media doesn't listen to us either. They say we would complain of misinformation. We see vaccine damage every day. People contact us all the time because of their vaccine damage.'{}'' (2022-06-14 15, 243 retweets)  
\end{displayquote}

Finally, we observe examples in which a larger number of issues are explicitly aligned:

\begin{displayquote}
    ``Never thought how fast people swallow absurdities like "weapons create peace". But clearly, mask was also freedom \#b2502 \#peace \#Corona \#propaganda'' (2023-02-25, 1350 retweets)\\[2mm]
  ``There are no women with penises. Climate change is not man-made. Unregulated migration is destroying our social system and endangering our everyday lives. Corona is an average disease. Vaccination is ineffective but not without side effects. With 70\% in taxes and levies, the state is taking almost all of the income from working people. Ukraine is not defending us or our “values”. Every now and then such banal truths have to be spoken in the face of the flood of disinformation.'' (2023-05-10, 1646 retweets)\\[2mm]
``How did people actually survive in the past, when there were no smart-ass recommendations for meat consumption, Karl Lauterbach wasn't panicking about the heat, nobody was constantly talking about CO2 and people were still allowed to decide for themselves which heating systems to install and which cars to buy?'' (2023-07-09, 129 retweets)
\end{displayquote}

These examples allow to generate hypotheses about certain meta-narratives that may guide the ideological narratives of the two camps. On the right, we observe a strong emphasis on individual freedom and reluctance to governmental interference. Furthermore, the skeptical narrative of the ``lying media'' and ``fear-mongering politicians'' transcends topical boundaries. On the left, we observe a recurring theme of solidarity and shared responsobility -- with the Ukrainian people, risk groups in Covid, the protesters in Lützerath -- that may serve as a larger ideological meta-narrative. 

These observations are a first approach at using the narrative lens to study how issues are aligned through emplotment. In the examples shown for the war in Ukraine, Covid and climate change, we observe a combination of repeatedly antagonized actors and direct analogies that may serve as strategies to align issues discursively and bundle them into coherent ideological sets.

\section{Discussion}
The present work has analyzed the discursive dimension of polarized debates through the lens of conflicting narratives. On a large dataset of tweets (approx. 20M) from trending topics in Germany between 2021 and 2023, where previous structural analyses have revealed strong polarization and issue alignment \citep{Pournaki2025}, this work investigates the discursive dimension by extracting the conflicting narratives of opposing opinion groups. Focusing on three strongly salient topics (the Russian invasion of Ukraine, the Covid pandemic and climate change), it demonstrated -- using the guided close reading method of extracting and reassembling narrative signals developed in \citep{Pournaki2024} -- how opposing opinion groups diverge in their interpretations of the same political events and issues. In particular, two ways in which narratives about the same political issue can be conflicting were identified: (i) they can invoke the same actors, but attribute different relationships to them, or, (ii) they can invoke different actors altogether, showing a divergence in the attribution of actantial importance within topics.

The first type of conflicting narrative is the one we largely observe in the case of the Russian invasion of Ukraine. The central actors (e.g. Ukraine, Russia, Germany, the USA, NATO, weapon supplies) are present in both narratives, but the role, in particular of NATO and the USA are diametrically opposite. While these actors want to stop the war in the left-leaning narrative, they are fueling it in the right-leaning narrative. Similarly, the elemental connection between human activity and climate change is disputed between the two sides in the issue of climate change. 

The second type of conflicting narratives, the omission and highlighting of different actors, can be also observed across the analyzed issues. In the narrative of Russia's invasion, the right invokes the actor Nord Stream 2, which is barely mentioned on the left. Similarly, vaccination side effects play a small role in the left-leaning narrative on the Covid pandemic, as does Bill Gates. Looking at the narratives on climate change, speed limits and the events in Lützerath play a marginal role in the right-leaning narrative, which instead focuses on alleged threats by the supermarket chain Lidl to the personal freedom to eat meat. These findings echo a central feature of political narratives: only elements that serve the purpose of ``fram[ing] [one's own] political positions as favorable'' \citep{Groth2019} are emplotted.

Analyzing the common narrative signals across the three topics, we find first evidence for the potential relationship between narratives and issue alignment. We observe a recurrence of certain antagonized actors and analogous relationships across the three topics we investigated (e.g. threats to Germany by certain political parties, threats to individual freedom by policies...). Connecting these findings to previous literature, we recall recent work by \citet{Lejano2020} analyze the narratives of climate change skeptics, where they observe a meta-narrative around the loss of individual freedom through climate policies. This is in line with the findings in our work, where we observe this pattern in the right-leaning camp for the topic of Covid (through measures) and Ukraine (Germans not being free to choose to intervene or not in the war). On the left, we observe signs of a meta-narrative around solidarity and shared responsibility that connects the three observed topics (solidarity with Ukrainian people, Covid risk groups, climate activists).

What remains to be investigated more thoroughly is how to model such broader meta-narrative patterns, thus allowing for a systematic analysis across a wider range of topics present in our dataset (see Tab.~\ref{tab:topicmodel} for all the topics present in the corpus). Such models could potentially build on structuralist approaches such as the one by \citet{Greimas1987} that would encompass the notion of ``genetic metanarratives'' hypothesized by \citet{Lejano2020}. Comparing climate skeptical narratives with anti-immigration narratives, the authors find the recurring pattern of an outside actor (``the eco-terrorist'', ``the foreigner'') threatening to transform a desired ``traditional way of life''. Further developing methods for the detection of existing, and the discovery of new such patterns can provide additional insights into the role of narrative in ideological polarization and issue alignment.

Aside from the potential for more comprehensive conceptual and computational models for metanarratives, the present work has limitations that call for future work. Firstly, the analysis of conflicting narrative networks assumes an internal coherence of narratives within the opinion groups. This must not necessarily be the case, especially for topics like the war in Ukraine, which has shown to divide the German Left party.
Furthermore, we recall here that we only investigated the links for which the communities' narratives differed -- looking at those parts of the narratives that are similar could provide a potential for mediation across camps. 
Additionally, narratives usually evolve over time, an aspect also left unconsidered so far. In future work, actantial networks can be analyzed over time to detect narrative shifts, for example through changes in link scores, or the introduction of new actors to the plot. This would allow deeper empirical insights into ``how narrative truth is fabricated, how the validity of narratives is constructed or deconstructed, and how diffferent modes of reception influences the felicity conditions of political narratives.'' \citep{Groth2019}.

Reflecting on the general appropriateness of actantial networks as representations of narratives, we may question whether the relationships they encode (supportive/conflictive/neutral) are complex enough to capture the potential nuances present in ideological narratives. While the full spectrum of verb actions (for instance, given by Verbatlas) is too extensive, there may be an intermediary level of verb categorization that could be envisaged in future work. 

Furthermore, additional work shall be done investigating the construction of in-group identities. This work has provided first steps towards this analysis by investigating nodes connected to a common ``we''. We identify several natural avenues of improvement: firstly, the pronoun ``we'' can be decomposed into its different deictic facets (inclusive, exclusive, general). Secondly, the extraction and labeling of actantial links can be sharpened by adding the dimension of the speaker: is the speaker part of the ``we'' that is evoked? This distinction would improve the direct interpretability of the actantial networks. Furthermore, more thorough analysis needs to be conducted to analyze the systematic antagonization of the out-group. While investigating the actantial ego network of the pronoun ``they'' could be a first approach, this would require to be complemented with coreference resolution to remove textual traces that do not correspond to out-group descriptions.

Reflecting on the representativity of the findings, we highlight that the present work has focused the analysis on one social media platform, Twitter/X, in a single country, Germany. Future work shall extend to different social media platforms and traditional media, as well as include different countries. The mathematical object of the actantial network lends itself to a systematic analysis of narrative diffusion through measures of graph similarity, which can yield insights into the role of different platforms and media sources play in distributing or influencing leading narratives around different issues.

To conclude, the observations presented in this work provide empirical descriptions of previously mentioned functions of narratives in conveying a political agenda \citep{Hartley2015,Groth2019}. By providing a lens through which political reality is interpreted, they reduce the complexity of political crises like war, pandemics, or climate change, by selecting specific events and actors to highlight, and assigning antagonist/protagonist roles to them. These findings suggest further evidence for the manifestation of an ``epistemological crisis'' postulated by \citet{Friedman2023}, where opposing groups diverge with respect to their interpretation of political reality. If both groups assume that their narrative is a ``true'' representation of reality, this can only lead to mistrust and antagonization of the out-group.
This work presented first attempts to empirically address these questions by providing a description of these diverging interpretive lenses in the form of conflicting narratives.

\section*{Acknowledgements}
The author thanks Eckehard Olbrich for fruitful discussions. The author acknowledges funding from the European Union's Horizon Europe framework (HORIZON-CL2-2022-DEMOCRACY-01-07) under project number 101094752 (\url{https://some4dem.eu}), and from the French government under management of Agence Nationale de la Recherche as part of the ``Investissements d'avenir'' program, reference ANR-19-P3IA-0001 (PRAIRIE 3IA Institute). 

\renewcommand\bibname{References}
\printbibliography[heading=bibintoc]
\newpage
\appendix
\section*{Appendix}
\renewcommand{\thesubsection}{A.\arabic{subsection}}
\renewcommand{\thefigure}{A\arabic{figure}}
\setcounter{figure}{0}
\renewcommand{\thetable}{A\arabic{table}}
\setcounter{table}{0}

\subsection{Topics in German Twitter trends 2021-2023}
\begin{table*}[ht]
\caption{Main issues discussed in German Twitter trends from March 2021 and July 2023 (taken from \citet{Pournaki2025}).}
\label{tab:topicmodel}
\begin{tabular}{lrrrrr}
\toprule
Topic & $N_{\mathrm{tweets}}$ & Retweet share & $N_{\mathrm{trends}}$ & $N_{\mathrm{trends}}$ with $|V| \geq 50$ & Polarized trends share \\
\midrule
German politics & 2276715 & 0.82 & 338 & 305 & 0.71 \\
Covid & 1862823 & 0.78 & 250 & 217 & 0.70 \\
Ukraine & 1701862 & 0.79 & 205 & 177 & 0.66 \\
Climate change & 1461830 & 0.80 & 157 & 133 & 0.60 \\
Sports & 920527 & 0.65 & 514 & 180 & 0.36 \\
Greetings and holidays & 822367 & 0.48 & 225 & 94 & 0.39 \\
Foreign politics & 768439 & 0.79 & 97 & 73 & 0.60 \\
Democracy & 709776 & 0.77 & 62 & 59 & 0.69 \\
Journalism/Media & 497964 & 0.82 & 50 & 44 & 0.64 \\
Police & 467491 & 0.85 & 61 & 55 & 0.65 \\
Social politics & 463404 & 0.77 & 73 & 58 & 0.52 \\
Social media & 416302 & 0.75 & 71 & 34 & 0.41 \\
Gender/LGBTQ & 408078 & 0.77 & 71 & 50 & 0.70 \\
Right wing extremism & 347301 & 0.82 & 17 & 16 & 0.81 \\
Energy & 318021 & 0.77 & 43 & 36 & 0.72 \\
Pop culture & 269714 & 0.59 & 175 & 55 & 0.31 \\
Racism & 244734 & 0.78 & 41 & 36 & 0.64 \\
Migration & 206589 & 0.83 & 17 & 15 & 0.60 \\
Religion & 185335 & 0.72 & 32 & 20 & 0.55 \\
Mobility & 138568 & 0.73 & 19 & 15 & 0.33 \\
Antisemitism & 137434 & 0.83 & 14 & 12 & 0.67 \\
Drug legalisation & 132207 & 0.74 & 21 & 16 & 0.25 \\
Gaming & 113756 & 0.65 & 64 & 9 & 0.11 \\
Music & 49365 & 0.55 & 60 & 14 & 0.21 \\
Abortion & 16567 & 0.76 & 4 & 3 & 0.67 \\
\bottomrule
\end{tabular}
\end{table*}

\subsection{Relation extraction prompt}
\label{sec:prompt}
We use the open weight model \texttt{Phi-4} \citep{Abdin2024} with the following prompt, where \{A1\} is replaced by the actor label of the agent, \{A2\} is replaced by the actor label of the patient, and \{tweet\} is replaced by the tweet to process:

\setlength{\fboxsep}{10pt}
\fbox{\parbox{\dimexpr\textwidth-2\fboxsep\relax}{\small \sffamily
You are an expert political analyst. In the following tweet, the author expresses a relation from "\{A1\}" to "\{A2\}". 
Provide a description of the relation in English in max. 3 words. Determine whether the relation is supportive, conflictive or neutral for "\{A2\}".\\

Definition of relation types:\\
Supportive relations include relations where "\{A1\}" approves of, causes, cares for, contributes to, helps, approves of, or creates "\{A2\}".\\
Conflictive relations include relations where "\{A1\}" disapproves of, attacks, betrays, prevents or reduces "\{A2\}".\\
A neutral relation applies when the connection is only evoked in reported or direct speech, or when the implied relation is neither supportive nor conflictive.\\

Do NOT determine the relation type based on general knowledge — only use what is stated in the sentence.\\

TWEET: \{tweet\}

}}\\

\subsection{Validation of actantial link labeling}
\label{sec:llm_validation}
We selected 100 tweets, along with the ARG0 and ARG1 actants extracted using the AMR approach, and hand-coded the implied relationship (supportive/conflictive/neutral) manually. We observed cases which are difficult to address, in particular because they raise the question of necessary context. Consider the following example, where the aim is to code the nature of the relationship between \textit{I} and \textit{masks}:

\begin{displayquote}
``Well, I will continue to wear masks.''
\end{displayquote}

As a human annotator, we are inclined to infer a supportive relationship for masks in general, since the act of wearing them, during the pandemic, symbolizes support for the measure. The LLM's response is neutral, with the following explanation:

\begin{displayquote}
The statement implies that the author will continue to wear masks. The relation between 'I' and 'masks' is neutral because the action of wearing does not inherently express approval or disapproval; it simply states a personal intention without additional context.  
\end{displayquote}

This explanation leads us to agree with the output of the LLM, since there are indeed no explicit textual cues that imply a strong support.  

There are clear cases, however, where the LLM fails to extract the correct relation type, in particular when the input text contains irony or sarcasm. This is something that could likely be mitigated with the increasing ability of future models. Consider the following tweet as an example, where we want to extract the relationship between ``we'' and ``Armin Laschet'':

\begin{displayquote}
``Yes, the streets are flooded; yes, these are extreme weather situations, but they have always existed! Just read the Bible. We have to elect Armin Laschet now, because only he can build us an ark!  \#HeavyRain \#LaschetWave'' 
\end{displayquote}

The LLM extracts a supportive stance since the tweet seems to praise Armin Laschet, while the expert human (especially knowing the political context) can infer the irony and detect the conflictive stance.

We compare the result from the LLM annotation to what we would get if we used a dictionary-based verb classification previously introduced in \citet{Pournaki2024} that we refer to here as a context-free dictionary approach (CFD). We report an agreement with the human annotation on the sample of 100 tweets that is significantly higher than the agreement with the CFD: the AMR+CFD approach yields an agreement of 46\% with the human annotation, while the AMR+LLM approach yields an agreement of 86\%. The low agreement with CFD may stem from different sources that warrant closer inspection. For one, the translation step from German to English could introduce discrepancies and erase linguistic nuances that are captured by the LLM through the possibility of processing the original text. Furthermore, the possibility to use the context, such as additional (sub-)sentences, instead of the verb alone, most likely leads to better predictions. However, we highlight that the output of the AMR-based extraction of \textit{actors}, which is fed to the LLM, is robust, even considering the translation. While additional validation and more thorough testing of various models is necessary, these preliminary insights expose the value of a hybrid approach combining AMR-based relation extraction with LLM-based relation labeling. 

\subsection{Close reading examples}
\label{sec:appendix:closereading}
In this section, we report a selection of translated tweets that were close-read based on the actantial networks in order to reconstruct the conflicting narratives described in Sec.~\ref{sec:results}.

\subsubsection{War in Ukraine}
\subsubsection*{Identity network}
We see that the left-leaning camp is strongly supportive of \node{Ukrainians}, the \node{defense} and \node{support} for Ukraine:

\begin{displayquote}
``On the way to \#Ukraine. Today in Kiev and tomorrow in Moscow, I will continue our talks on the continuing very serious situation on the border of Ukraine. In Kiev, it is important for me to express our continued solidarity and support to Ukraine. (1/2)'' (2022-02-14, 327 retweets)\\[2mm]
``We are thinking about the people in \#Ukraine. And those who are now on the run. We are looking for people who can translate pressing health information. It is about being able to give refugees timely information in their hands. Thank you! Please RT.'' (2022-02-28, 491 retweets)
\end{displayquote}

The (Russian) \node{attack} is condemned, as well as \node{demonstrations}, which refers to the ``demonstrations for peace'':

\begin{displayquote}
``Maybe we could agree to stop talking about a "peace demonstration" today? \#b2502 \#StandWithUkraine'' (2023-02-25, 568 retweets)\\[2mm]
``The mass graves in \#Lyman leave you behind speechless. 340 people were hidden anonymously by the \#invasion troops - including 60 children. The \#Ukraine has now properly buried them and is trying to clarify the identities via DNA. Let's stop \#Putin.'' (2023-01-07, 558 retweets)\\[2mm]
``We should not only condemn the Russian air attacks on peaceful Ukrainian cities, civilians and infrastructure, but also speed up delivery of air defence and now deliver the weapons that have not yet been delivered - firearms and combat tanks.'' (2022-10-10, 486 retweets)  
\end{displayquote}

Turning to the nodes specific to the right-leaning \node{we}, we observe a supportive relation towards \node{freedom} and \node{Nord Stream2}:

\begin{displayquote}
  ``"They can and should play an important role as honest brokers. The prerequisite is that we draw the right consequences and rebuild lost trust, sovereignty and freedom of action," says Alice \#Weidel to the \#Attack War on Ukraine. \#Bundestag'' (2022-02-27, 433 retweets)\\[2mm]
``We should not do what the perpetrators want \& forget NordStream 1 \& 2. On the contrary. What we should demand is the immediate repair \& the restarting of the two pipelines. Right now. Everything else would be in the sense of the perpetrators \& would harm us massively.'' (2022-09-28, 762 retweets)
\end{displayquote}
The \node{NATO} and \node{Europe} are antagonized:
\begin{displayquote}
  ``Zelensky: “The US will have to send their sons and daughters just as we send our sons and daughters to war. And they will have to fight because we are talking about NATO, and they will die.”'' (2023-03-01, 1194 retweets)\\[2mm]
``We should not leave Europe to the 'Leyens'.''\footnote{This is a play on words. The German word ``Laie'' means ``amateur'', the tweet implies that Ursula von der Leyen is an amateur unit to lead the European Union.} (2022-09-16, 287 retweets)  
\end{displayquote}

Moving to the nodes that are connected to both camps shows actors on which the stances differ. Looking at \node{gas}, the right-leaning camp is denouncing a double standard connected to stopping the gas sales with Russia:

\begin{displayquote}
  ``We do not want gas from Russia because they launched an "attack war" against Ukraine. Instead, we are now buying gas from Azerbaijan. They just launched an attack war against Armenia. The right attitude is so important.'' (2022-09-13, 634 retweets)\\[2mm]
``Gabor Steingart at \#Maischberger: We have paid off the gas ourselves with the sanctions against Russia. This is an economic war against our own (German) population, which we cannot finance. I agree! Who else?'' (2022-09-20, 611 retweets)  
\end{displayquote}

On the left side, the issue of \node{gas} is connected to Russia not meeting contracts:

\begin{displayquote}
``Once again for stupid: We don't get gas because \#Russia doesn't meet its contracts! It doesn't matter how many pipelines don't get gas. \#NordStream2'' (2022-09-27, 729 retweets)
\end{displayquote}

Furthermore, we observe that the stance towards \node{weapon} supplies is different on the left and the right. The right side is vocally against the supply of weapons:

\begin{displayquote}
``It is precisely this point that worries me: if we deliver weapons - what happens to them later? No one can check where they are and, unfortunately, in Afghanistan we see that things can go wrong...'' (2022-06-03, 724 retweets)\\[2mm]
``As an opposition they promised: "No weapons in war zones!" As a government: "We need weapons only in war zones." As an opposition they said: "Climate has priority! Quit fossil energies!" As a government: "We urgently need coal." Greens even deceive the climate.'' (2023-01-24, 540 retweets)  
\end{displayquote}

On the left side, there is a positive stance towards \node{weapon} supplies:

\begin{displayquote}
  ``Russia is waging its war of destruction from a safe distance. Where is our answer? By refusing Ukraine extensive weapons, fighters + tanks, we are forcing it into an asymmetric war. That is politically, militaryly and morally wrong. « Gerashchenko\_en: A woman curses Russians. It seems she lost her son in the attack. Her pain is unbelievable. Please watch the video till the end (I know it's not the easiest one to watch). Look at the scale of destruction. And look at the number of people helping. — »'' (2023-01-14, 325 retweets)\\[2mm]
``Germany must stop hiding behind the USA. We have tanks \& can deliver. We must finally do so now, so that this brutal war that is taking place in our neighbourhood ends as quickly as possible. \#Ukraine'' (2022-09-12, 674 retweets)    
\end{displayquote}

As for \node{war} and \node{peace}, there are conflicting stances as well. The left is not calling for an entering of Germany into the war, but for support of Ukraine:

\begin{displayquote}
  ``Russia is threatening to dry up Nord Stream 1. The European response can only be: it should. If, for good reasons, we are not prepared to intervene directly in the war, then at least we must be prepared to sacrifice part of our prosperity.'' (2022-03-07, 658 retweets)\\[2mm]
``If Ukraine stops fighting, Ukraine ends. If Russia stops fighting, this war ends. And that is why we support Ukraine, because we want this war to end. And because we want real peace. \#StandWithUkraine'' (2023-02-24, 373 retweets)\\[2mm]
``The bravery of the Ukrainians must be saluted. Every day we must examine whether we can do more to help them in this war. \#Ukraine must win this war. CL'' (2022-09-12, 276 retweets)  
\end{displayquote}

The right, on the other side, calls for \node{peace} by not escalating the war and letting it run its course:

\begin{displayquote}
  ``\#Muetzenich is right: \#StrackZimmermann \& others are irresponsibly leading us into direct war participation. Anyone who demands delivery of heavy \#combat tanks today will call for aircraft or troops tomorrow.'' (2023-01-22, 1652 retweets)\\[2mm]
  ``We all don’t want a war! Stop agitating against Russians! Let’s not get back into a useless war that only benefits the powerful! The media lie to us, the politicians lie to us, listen to your heart!'' (2022-03-02, 519 retweets)\\[2mm]
``No, damn it. For \#Ukraine neither arms should be supplied nor soldiers sacrificed. As Helmut Schmidt said, “It’s not our business.” Would the Ukrainians fight for Germany? No! We should not participate in war for this corrupt country.'' (2022-02-27, 389 retweets)  
\end{displayquote}

\subsubsection*{Conflict network}
There is a conflictive link from \node{Ukraine} to \node{Europe} on the right camp:

\begin{displayquote}
``The United States has been preparing Ukraine for this war for 8 years in order to destroy Europe economically, financially and militarily through the back door. We are celebrating our own demise because someone has called "Putin is evil".'' (2022-05-14, 324 retweets)  
\end{displayquote}

We observe traces of antagonization on both sides. On the left side, the \node{AfD} is detrimental for \node{Germany}:

\begin{displayquote}
``The AfD is so “patriotic” that it would donate Germany for a knob to Putin.'' (2022-03-02, 565 retweets)  
\end{displayquote}

while the \node{Green Party} are the main antagonist harming \node{Germany} on the right:

\begin{displayquote}
``2 years ago a lot of people still laughed about this action! Today, the `Ampel' is not 2 years in office, nobody laughs anymore. At record speed the country is ruined and society is divided and inflamed against each other! The Greens are the biggest danger for Germany!'' (2023-02-28, 491 retweets)  
\end{displayquote}

A central conflict is attached to the role attributed to \node{NATO} and the \node{USA} in the war in Ukraine. On the left side, the \node{USA} and \node{NATO} are helping \node{Ukraine} and are trying to end the \node{war}:

\begin{displayquote}
``Have \#Wagenknecht/s “justification” with the alleged statement of the \#Rand Corporation advised them to \#start negotiations urgently because the \#Ukrainian war could not be won reviewed because it didn't seem plausible to me. Result: \#FakeNews (see sources in the... « SandraNavidi: Does anyone actually review \#Wagenknecht's claims? Rand: It will be important for America to persevere in supporting Ukraine for the long haul ... Stability in Europe is a vital US national security interest that may be served best thru stopping \#Putin — »'' (2023-02-21, 188 retweets)\\[2mm]
``The perhaps most important political visit in the history of the independent Ukraine. Joe Biden surprised \& impressed the Ukrainians with it. His message: The USA will support Ukraine as long as necessary. Arrangements for the @Dayshow.'' (2023-02-20, 108 retweets)  
\end{displayquote}

On the right side, the \node{USA} and \node{NATO} are profiting from the \node{war} and don't want the it to stop:

\begin{displayquote}
  ``Putin had offered the Americans to negotiate with Russia on the status of Ukraine. Joe Biden could have easily prevented this war. But the US wanted this war and have rejected Putin, at the expense of Ukraine and Europe.'' (2022-09-04, 808 retweets)\\[2mm]
  ``What is your opinion? Has the @USkraine deliberately rocketed Poland in order to provoke the total war between NATO and Russia with a \#FalseFlag action?'' (2022-11-17, 114 retweets)\\[2mm]
``\#NATO expansion to the east contradicted Western promises in the 1990s. In order to prevent escalation of the conflict in \#Ukraine, Russian security interests must also be taken into account \& a NATO accession of Ukraine must be excluded. \#AnneWill'' (2022-02-21, 581 retweets)  
\end{displayquote}

We see an interesting reversal of situations between \node{media} and \node{Russia}. On the left side, \node{Russia} is being accused of manipulating the \node{media}:

\begin{displayquote}
``The Russian strategy: to turn off the free media quickly, then to turn on their own propaganda and thus take control. That way then to organize "elections" and use a pro-Russian government. Like in Crimea.'' (2022-03-01, 197 retweets)  
\end{displayquote}

while on the right, the \node{media} is against \node{Russia}:

\begin{displayquote}
``1/2 The kerosene lie: Currently, the media is spreading the lie that M1 Abrams needs kerosene, so it is not suitable for that because of supply problems. His gas turbine can use EVERY burning liquid as fuel in any mixture:'' (2023-01-21, 111 retweets)  
\end{displayquote}

On the right, the \node{USA} are allegedly attacking \node{Nord Stream2}:

\begin{displayquote}
``The \#USA torpedoes \#NordStream2. At the same time, \#Russia becomes the third largest oil supplier for the US itself – thanks to sanctions.'' (2022-02-23, 382 retweets)  
\end{displayquote}

An interesting distinction is also observed between the left and the right with respect to \node{freedom}, an actant typically invoked within the right-leaning narrative. On the left, \node{Ukraine} is defending ``our'' \node{freedom}:

\begin{displayquote}
``Since 1989/90, German politics has hardly changed as fundamentally as it does today. It was time. \#Ukraine is now defending our freedom as well.'' (2022-02-26, 17 retweets)  
\end{displayquote}

While the right claims that this is not ``our'' \node{freedom} that is being defended:

\begin{displayquote}
``Friedrich Merz claims that Ukraine also defends "our freedom". What's with these lies? No German has asked Ukraine to fight for him. We should not be talked into a war that is not ours! \#DasIstNichtMeinKrieg'' (2022-05-08, 357 retweets)
\end{displayquote}

A central allegation that separates the left and the right is whether and under what conditions \node{Ukraine} seeks \node{peace}. The left is clear that \node{peace} can only be had if \node{Ukraine} remains sovereign:

\begin{displayquote}
``It is a strawman argument, because -- of course wanting peace is per se good. Ukraine would also like peace - AND wants to continue to exist. The latter will be taken away from it if it subjugates to Russia.'' (2023-02-11, 24 retweets)
\end{displayquote}

The right-leaning narrative on that issue is not that clear and insinuates that \node{Ukraine} is not that interested in \node{peace}, but rather profits off of other countries:

\begin{displayquote}
``Billions of euros go to other countries. With billions we are supporting the Ukraine war and the refugees enjoy in Germany "all inclusive" holidays. Now the Green leader said that the kindergarten standards are no longer financable. Future group size now with 30 children'' (2022-09-13, 67 retweets)\\[2mm]  
``The \#Ukraine demands to join the war, joining NATO and the EU, and money. Freely available money. Why does Ukraine not actually demand peace?'' (2022-02-27, 12 retweets)  
\end{displayquote}

There is a different evaluation of \node{Vladimir Putin} and \node{children}. On the left, the ICC court order is reported:
\begin{displayquote}
``The International Criminal Court has issued arrest warrants against Putin for the illegal deportation of children from \#Ukraine to Russia. Now \#Putin can no longer travel freely without having to fear being arrested. Important sign of solidarity \#PIRATEN'' (2023-03-17, 37 retweets)  
\end{displayquote}
On the right, he is considered a hero for saving children from war zones:
\begin{displayquote}
``"The Internal Criminal Court, which the \#US does not recognize, has issued a \#warrant against \#Putin for having \#evacuated \#children from a \#war zone that is being bombarded with \#US weapons, which have killed over 20 million people in 37 countries since the Second World War...'' (2023-03-18, 525 retweets)  
\end{displayquote}

Finally, there is the strong mismatch between the left and the right with respect to the role of \node{weapon supplies} to end the \node{war}: while the left claims that the \node{supply} needs to be made to end it, the right claims that the \node{supply} will just escalate the \node{war}.

\begin{displayquote}
``"Weapons prolong the war" often comes as an argument from self-explained pacifists. The opposite is the case. How many more cities could have been liberated if Ukraine had already received combat and artillery tanks - also from Germany - in the summer?'' (2023-01-23, 80 retweets)  
\end{displayquote}

On the right, we see an example that leads us to the next issue:

\begin{displayquote}
``Never thought how fast people swallow absurdities like "weapons create peace". But clearly, mask was also freedom \#b2502 \#peace \#Corona \#propaganda'' (2023-02-25, 1350 retweets)
\end{displayquote}

\subsubsection{Covid}

\subsubsection*{Identity narrative}
We observe different stances on \node{testing}:

\begin{displayquote}
 ``For people who visit relatives in hospital or care, for example, or even care for relatives, we must offer reliable free tests for COVID. Christmas time must not be a risk for these people.'' (2022-11-22, 151 retweets, left cluster)\\[2mm]
``Our government boycotts the constitution. We boycott the tests \#Testboycott'' (2021-08-10, 61 retweets, right cluster)
\end{displayquote}

Looking at the nodes that are uniquely connected to one or the other camp, we find that the right is focused on \node{side effects}, freedom.
Here we highlight another actor, which is \node{data}.

\begin{displayquote}
``So one wonders: Why don’t we have these data in Germany? [link to article claiming that 80\% of Covid deaths are not due to Covid but other diseases]'' (2021-08-31, 349 retweets)
\end{displayquote}

On the left, one is concerned with protecting the \node{population}, and protecting \node{children} by keeping \node{schools} closed:

\begin{displayquote}
``This video is worth seeing. It shows correctly why the concerns of the vaccinators are unfounded. Why the video did not exist before does not matter. We are moving forward towards protecting the population.'' (2021-12-19, 1634 retweets)\\[2mm]
``Early 2020: If we only had the fastest \& a vaccination, then we could end the pandemic Early 2021 (fastest \& several vaccines available): Oh, let's just run it \& check young age groups \& children \#SARSCoV2 \#COVID19 \#YesToNoCovid'' (2021-04-12, 974 retweets)\\[2mm]
``“The first thing we need to do is to open the schools.” Of course, the first thing I am going to do is to open the places that are worst protected, where about 10 million people come together for very long periods of time, most of whom are likely to carry the virus unnoticed. That makes sense.'' (2021-01-01, 423 retweets)
\end{displayquote}

An interesting distinction can be observed for the common links towards \node{life}. In the left-leaning narrative, the main tweets revolve around \textit{saving} lives:

\begin{displayquote}
``Since 379 days!!!! we, the event industry, are in the real \#harterLockdown! We also want to work again. We also want to save lives! So now: \#harterLockdownNow \#SchulenKitasBuerosZu \#LebenRetten \#YesToNoCovid \#DoItKanzlerin \#AlarmstufeRot'' (2021-04-05, 305 retweets)\\[2mm]
``So that we can get the pandemic under control quickly, protect health, lives and economies and meet and celebrate again in pubs and concerts in the summer ... \#harterLockdownNow'' (2021-03-28, 241 retweets)\\[2mm]
``I'm a mother and I'm calling for a \#hardLockdownNow to be put in place so that we can save lives!'' (2021-03-28, 124 retweets)
\end{displayquote}

On the right, we rather observe a call for going back to the pre-pandemic \node{life} without government-imposed restrictions:

\begin{displayquote}
``The \#Corona incidence in Germany has now risen to 1500 and nobody notices anything about it. I remember that with an incidence of 50 we have completely turned off all public life. \#ItsoverKarl'' (2022-03-14, 568 retweets)\\[2mm]
``With the \#Lockdown \& other measures, many believe we are saving lives. Children have been forgotten. How will we respond to them when they ask later, what have we done to protect them? Strengthen your immune system? Strengthen your life?'' (2021-10-31, 25 retweets)
\end{displayquote}

The \node{vaccination} is criticized heavily by the right-leaning camp:

\begin{displayquote}
  ``After 3 years of propaganda it is obvious: - Covid was just a flu - The mortality rates were fake - Masks were useless - The vaccination was useless and harmful - Closing schools was useless - The high priests of the Covid cult are corrupt businessmen What now?'' (2023-02-11, 4048 retweets)\\[2mm]
``Seriously, what's going on here? The pandemic is over. SARS-CoV-2 is endemic. Omikron usually has only cold potential. Hospitals are not overloaded. Nevertheless, there is still a debate about vaccination, masking, etc. irrational measures. It's a mass psychosis!'' (2022-02-22, 2841 retweets)
\end{displayquote}

In this context, we observe a focus on \node{side-effects}, which is important in the right-leaning camp, but not on the left:

\begin{displayquote}
  ``1. When I noticed in 2020 that the R value had fallen under one, it disappeared from the discussion. 2. When the vaccination effectiveness became negative, the RKI submitted the reports. 3. When the side effects rose, the PEI took the database from the network. Do you trust the government?'' (2022-07-07, 3610 retweets)\\[2mm]
  ``Many acquaintances got the vaccine only for convenience. They were allowed to go to the restaurant. I wasn't. They were allowed to travel. I wasn't. They were allowed to go to Christmas market. I wasn't. Now even BMG and ÖRR are admitting side effects and they get to deal with the fear. I didn't.'' (2022-08-13, 2369 retweets)
\end{displayquote}

Conversely, the \node{vaccination}'s effectiveness is highlighted on the left:

\begin{displayquote}
``A global pandemic. 4.5 million people die. There is a vaccine. But only in rich countries. Free. Voluntary. Nevertheless, people protest against the vaccination. And against tests. And against masks. And against restrictions in public spaces.'' (2021-09-04, 2780 retweets)\\[2mm]  
``(1) This graph from Israel shows what lies before us . It is quite clear that covid vaccination will significantly reduce both the incidence and mortality when more than 50\% of the population has first vaccination. We will achieve this by the end of May. So there are still 6 weeks left. What do we do?'' (2021-04-18, 1618 retweets)
\end{displayquote}

\node{Lockdowns} are discussed almost equally much in both camps, but the position towards them is a different one. While the left largely approves of them:

\begin{displayquote}
  ``The Federal Health Minister calls for immediate lockdown. Karl Lauterbach calls for immediate lockdown. Scientists call for immediate lockdown. Association of Intensive Medicine calls for immediate lockdown. WHOSE ADVICE ARE YOU WAITING FOR? BATMAN?! \#COVID19 \#hardLockdownNow'' (2021-03-28, 4019 retweets)\\[2mm]
  ``(1) There must be a nationwide lockdown as soon as possible. 30,000 newly infected and 600 dead in one day (!) without any improvement in sight. So we cannot celebrate Christmas. Nothing is Christian about shopping and celebrating before we act'' (2020-12-11, 1069 retweets)\\[2mm]
``The AfD is - against mask duty - against lockdown - against test duty - to ensure that positive tests are not considered to be infected The AfD reacts to Corona exactly as it does to the climate crisis: simply deny and do nothing.'' (2021-04-10, 1013 retweets)  
\end{displayquote}

The right tends to condemn the \node{lockdown} measures:

\begin{displayquote}
``Unvaccinated in Poland: hotel, restaurant, shopping, swimming pool. Everything possible. Unvaccinated in Germany: lockdown, exclusion and confinement to private. Same virus, similar incidence and only 54\% vaccination rate in Poland. Tell me! Germany is finished. \#Unvaccinated'' (2021-12-04, 1205 retweets)\\[2mm]
``With the prospect of an end to the measures, the citizens swallowed lockdown, masking, exclusion and vaccination. But the politics did not keep their promises. It remains endless and unfortunately also evidenceless! With @Karl\_Lauterbach it will start again from the front in autumn.'' (2022-06-02, 1199 retweets)\\[2mm]
``++EIL++ @Karl\_Lauterbach has blown up the evaluation of the measures according to the WORLD. Should fall lockdown, 2G/3G, masking force in D therefore come again without any evidence, illegal and internationally isolated?'' (2022-06-02, 1181 retweets)
\end{displayquote}

\subsubsection*{Conflict network}
On the right, there are a large number of connections that do not exist on the left at all. For one, there is the connection between the \node{pandemic} and \node{Bill Gates}:

\begin{displayquote}
``Bill Gates is already planning the next pandemic... In India the production is very cost-effective and the value creation is probably the biggest. The virus of the "next pandemic" is certainly already sleeping in the laboratories. The vaccine as well. It is simply still grotesque what we are told!'' (2021-09-15, 143 retweets)
\end{displayquote}

We find the attack on \node{freedom} as well, by \node{politicians} and other entities:

\begin{displayquote}
``Some politicians want to gradually downgrade freedom and civil rights. This totalitarian-inspired \#Scheinfürsorge is clearly explained \& is already openly debated. Only incalculable scream for a guardian, but not self-thinking citizens! $=>$No vaccination obligation'' (2022-02-08, 133 retweets)\\[2mm]
``If I read politicians talking about \# vaccination obligations, "more freedoms" for \# vaccinated people (so no more fundamental rights), etc. here just before the elections... ...then I don't believe in the elections anymore!'' (2021-07-26, 65 retweets)
\end{displayquote}

\node{Politicians} instate \node{fear}:

\begin{displayquote}
``100 hospitals were closed during the pandemic and 4,000 beds were dismantled. At the same time, politicians stand in front of the camera and make people fear that there will be too few beds. Yes, now..'' (2021-11-11, 554 retweets)
\end{displayquote}

On the left, \node{Jens Spahn} is criticized for not delivering the \node{vaccine} in time:

\begin{displayquote}
``Shortly before the EMA release of the bionic vaccine for children aged 5-12 Jens \#Spahn contains this vaccine and supplies primarily only the vaccine from Moderna. He justifies it by the fact that Moderna could fail. In my opinion, it serves rather the purpose >> 1/x'' (2021-11-20, 798 retweets)
\end{displayquote}

\node{Karl Lauterbach} is a central antagonist in the right-leaning narrative:

\begin{displayquote}
``If Karl Lauterbach says that unvaccinated caregivers would have done nothing to cope with the pandemic and would not have the right to demonstrate, what kind of human image does this man have? If that is not enough for a resignation, then what? \#lauterbachmussweg'' (2022-06-23, 615 retweets)\\[2mm]
``Lauterbach is already back in panic mode and is screaming for masking, vaccination and other measures. A country in the grip of an allegedly ill man. How long can this be allowed to last?'' (2022-06-12, 618 retweets)\\[2mm]
``Lauterbach in the Bundestag: The whole country will be imprisoned without vaccination. I speak for millions of Germans when I say: This country is the only one imprisoned by Karl Lauterbach. And this can only be ended by flying out of office!'' (2022-03-17, 886 retweets)\\[2mm]
``Lauterbach scolded nurses and doctors: "Nothing contributed to combating the pandemic".'' (2022-06-22, 215 retweets)
\end{displayquote}

\node{Science} is invoked by the right, again, implying that there is a ``real'' science:

\begin{displayquote}
``Completely clear, \#immunity obligation There is no scientific reason for an immunity obligation. It doesn't matter what @Karl\_Lauterbach or other politicians say. Politically, the immunity obligation may be desired - it is not meaningful.'' (2022-03-17, 508 retweets)  
\end{displayquote}

While on the left, there is a critique of the entanglement between \node{science} and \node{politics}:

\begin{displayquote}
``The RKI has the task of making scientific recommendations for political action - also in public. Scholz and Lauterbach must not be allowed to dictate to the RKI. Science must be independent of politics.'' (2021-12-21, 236 retweets)  
\end{displayquote}

\node{Christian Drosten} is also antagonized on the right:

\begin{displayquote}
``Drosten has already committed suicide in 2009 with his misjudgments to the population. Many who believed in him are still suffering from health damage. At the first demos in April 2020, several were represented. They feared further harm early on.'' (2021-12-30, 225 retweets)  
\end{displayquote}

One central element of division is the role of \node{masks} in preventing the virus spread. The left tends to strongly support the use of masks:

\begin{displayquote}
``Superspreading outbreaks prove: the virus can stay for hours in the \#air of indoor rooms. 1.5 - 2.0 \#distance does not protect against contaminated \#aerosols in toilets, restaurants or classrooms. Safe \#masks protect non-immunized, recovered and infected \#COVIDisAirborne'' (2021-10-09, 241 retweets)
\end{displayquote}

Meanwhile, the right is not that convinced:

\begin{displayquote}
  ``I would like to embarrassingly point out once again that the effectiveness of \#masks against viruses is a pure invention of politics. \#MaskeAb \#DiviGate \#Freedom'' (2021-05-17, 104 retweets)\\[2mm]
``Mainstream media and \#Ethicrat are only coming on now: masks hurt children. Parents, teachers, judges and experts who have been saying this for two years are being persecuted. Are there any more house searches, Minister @GruenerDirk?'' (2022-04-05, 764 retweets)\\[2mm]  
``1. Twice vaccinated must wear masks because they are as contagious as non-immune ones. 2. Twice vaccinated must not be tested because they are not contagious. Any court that accepts this nonsense would be a VOLKSGERICHTSHOF. Sue them! 1/2'' (2021-08-12, 540 retweets)
\end{displayquote}

There is a fundamental distrust in the \node{government} in the right-leaning camp, connected to safety of the \node{vaccination}:

\begin{displayquote}
``Due to the fact that I am not vaccinated, I am 100\% safe from side effects and 99.8\% safe from Covid. I think my chances of surviving the pandemic are very good. Nevertheless, thank you for the charming "offer". \string^\string^ \#BoosterImpfung'' (2021-11-04, 754 retweets)
\end{displayquote}

We find a positive connection on the right from \node{politicians} to \node{fear}, especially Green politicians:

\begin{displayquote}
``The Greens want to maintain their policy of fear, because fear eats freedom. And with free people, these composted activists of the Prohibition Party can't do nothing!'' (2022-12-27, 395 retweets)
\end{displayquote}

We observe the connection between \node{children} and \node{Long Covid}, on the left:

\begin{displayquote}
``I wonder why we all act as if it were normal, even though we have a pandemic? Almost 1,000 people die a week, the trend is rising. Children infect schools and carry the infections on. Children get sick and get LongCovid. Normal?!'' (2021-11-09, 180 retweets)\\[2mm]
``\#Streeck spreads scientifically and statistically refuted falsehoods in \#Lanz: neither are children infected with \#Omikron so often (as the high incidences in schools show), nor is an infection harmless for children. Not to mention \#LongCovidKids.'' (2022-01-06, 164 retweets)
\end{displayquote}

Furthermore, \node{schools} are considered drivers of the \node{pandemic} on the left:

\begin{displayquote}
``Largest drivers of the pandemic: schools Second largest drivers: workplaces, businesses Exactly the two areas that are the focus of our measures. (Analysis of the sources of infection in Belgium)'' (2021-03-26, 658 retweets)
\end{displayquote}

\subsubsection{Climate change}
\subsubsection*{Identity narrative}
On the left, there are positive relations to the concept of \node{limit} (here referring to both the 1,5 degree limit and speed limits on highways), and the Lützerath \node{protests}:

\begin{displayquote}
  ``[Breaking!] While over 600,000 people were on the \#climate strike, RWE has today prepared the demolition of the first farm in Lützerath. Come on September 29th! Here we defend the 1.5° limit! \#LützerathBleibt!'' (2021-09-24, 314 retweets)\\[2mm]
  ``Transport and Climate Experts: We do not need new highways. What we need is a paradigm shift in terms of motorised individual traffic, a speed limit, new railways and an expansion of public transport. Government: New highways? Come now!'' (2023-03-29, 53 retweets)\\[2mm]
``The profit interests of RWE are put before the climate protection goals. That is why we are here and participate in the protest. Tonight I arrived in \#Lutzerath and stay here in the camp. \#luetzi remains \#LützerathUnräumbar \#dielinke With @voglerk'' (2023-01-09, 175 retweets)  
\end{displayquote}

A critical position is taken towards \node{cars}, \node{coal}, and the lax \node{climate policy} of the government:

\begin{displayquote}
  ``While we still put cars above people in this country, the Netherlands has understood in such a wonderful way that a city must not be oriented towards the needs of the strongest, but towards the needs of the weakest fellow citizens.[1]'' (2023-03-08, 294 retweets)\\[2mm]
  ``Nothing for weak nerves - but unfortunately such pictures are also necessary for rethinking: Not only humans, but also (use) animals die of heat, currently 1000 times in Kansas/USA. This is only the beginning of a development that will stop when we stop with coal, oil \& gas « rawsalerts: \#BREAKING: Thousands of cattle died in Kansas due to Dangerous Heat \#Kansas | \#USA More than 20 states are seeing dangerously hot temperatures impacting nearly 100 million Americans as Reports of Thousands of cattle have died in Kansas due to a dangerous temperatures — »'' (2022-06-16, 427 retweets)\\[2mm]
``Reminder! So gigantic was the \#climate strike with 1.4 million people - exactly 2 years ago: when the GroKo passed its anti-constitutional climate package. Now we can finish off with this catastrophic climate policy. Let's all remember it today! From the strike \#AlleFürsKlima'' (2021-09-24, 175 retweets)
\end{displayquote}

Looking at the nodes connected to both camps, we see that \node{violence} is described on both sides in relation to the protests in Lützerath. The right-leaning cluster tries to deligitmize the protests:

\begin{displayquote}
``For 5 days we have been researching in \#Luetzerath. There, the media and politics are stylizing a radical climate camp for resistance. We have spoken to activists and documented violence. Many never talked about the climate, but about systemic turmoil and militancy. Saturday, 12:00 pm the film.'' (2023-01-12, 261 retweets)
\end{displayquote}

Meanwhile, the left warns of actual \node{violence} in relation to climate change:

\begin{displayquote}
``A slight taste of what people will become if there is not enough water and food for everyone due to the deliberately escalating climate crisis. We are here to turn away violence and protect civilization. What are you doing, @Chancellor? [embedded video of car owners dragging climate activists off the street]'' (2023-05-23, 641 retweets)  
\end{displayquote}

The left-leaning camp warns of the dangers of \node{CO\textsubscript{2}} emissions:

\begin{displayquote}
``Our CO2 emissions are deadly. Nature magazine has computed this exactly. 4434 tons of CO2 cost a human life. Under Lützerath lie 280 million tons of coal. RWE wants to burn it. That's 63148 deaths. \#LuetzerathUnraeumbar'' (2023-01-08, 131 retweets)
\end{displayquote}

The right claims that the dangers related to \node{CO\textsubscript{2}} are not that serious:
\begin{displayquote}
``So let's be damn foolish and ask ourselves: How many of the victims of the \#FloodCatastrophe could still live if we had put just a quarter of the crazy means that we use to reduce CO2 in vain into flood-prone areas?'' (2021-07-16, 97 retweets)
\end{displayquote}

Similarly, we see that the \node{climate crisis} is played down on the right.
Forest fires, for instance, are not really increasing in frequency, and if they are, it is due to arson instead of global warming:

\begin{displayquote}
``The longest European time series for forest fires exists for the southern EU Member States. This shows a trend towards *less* forest fires and *less* burnt land. @GoeringEckardt Let us fight disinformation together. « GoeringEckardt: Even in Europe, more and more forest fires are raging. In the summer of 2022, fires destroyed more European forests than ever before. We need to fight the climate crisis consistently, also to protect our own health! — »'' (2023-06-08, 425 retweets)\\[2mm]
``"Even in Europe, more and more forest fires are raging" Why are you lying, Mrs Göring-Eckardt? Especially since every forest fire goes back to arson. « GoeringEckardt: Even in Europe, more and more forest fires are raging. In the summer of 2022, fires destroyed more European forests than ever before. We need to fight the climate crisis consistently, also to protect our own health! — »'' (2023-06-08, 260 retweets)
\end{displayquote}

Meanwhile, on the left side, the \node{climate crisis} is a real threat:

\begin{displayquote}
``Carla \#Reemtsma is being attacked here on Twitter - because she said at \#hartaberfair what should have been clear for a long time: that we need to act now to stop the climate crisis. Share in solidarity with Carla now! « Campact: Over 50 billion annually flow into fossil subsidies. Individuals cannot end the climate crisis. We need political measures! »'' (2021-11-02, 142 retweets)\\[2mm]
``At the current temperatures cooling by shade, fan, air conditioning and outdoor swimming pool is high in the course. However, in the case of heat we must also talk about the climate crisis above all. And so that it does not escalate even more, there must finally be an end to fossil subsidies!'' (2022-06-18, 138 retweets)  
\end{displayquote}

Looking at the actors only connected to the right-leaning cluster, we observe conflictive relations towards \node{``climate terrorists''} and \node{insects}:

\begin{displayquote}
  ``Throwing Molotov cocktails at policemen impressively proves that the ``Unwort'' of the year is rather a sad reality\footnote{Every year, a panel of German-speaking linguists critically selects the ``Un-word of the year'', a popular term that violates factual appropriateness or principles of humanity. In 2022, this word was ``climate terrorists''}. Yes, we also have to do with \#ClimateTerrorists. A result of fatal politics by the way.'' (2023-01-11, 325 retweets)\\[2mm]
``Why should we eat insects? Where are they produced? I don't know any insect manufacturing company! What is the CO2 balance of insect production? What if eating causes diseases? There is no experience with mass consumption of insects!'' (2023-01-18, 103 retweets)  
\end{displayquote}

The edge towards \node{meat} is conflictive because of the fears of end of consumption:

\begin{displayquote} 
``Formation is continuing to take off! The worst is that the companies are doing it in advance obedience! Discounter \#Lidl wants to educate us to renounce meat!'' (2023-02-02, 210 retweets) 
\end{displayquote}

\subsubsection*{Conflict network}
We see a distinction in the role \node{Germany} plays in mitigating \node{climate change}. On the right, we see a sub-narrative of ``response skepticism'' in which Germany should not act because it will not make a difference considering the global scale of the problem:

\begin{displayquote}
  ``In the end, a \#climate strike is just big nonsense. We can make the whole of Germany climate neutral, but it will not have any effect, because nobody in the world will follow our example. Therefore, go to work or to school.'' (2021-09-24, 120 retweets)\\[2mm]
``Do these dreamers believe that climate change will stop at the German borders if Germany becomes climate neutral, but nobody else will stick to it? By the way, our uncontrolled migration policy will cost us many times as much as the 900 billion in the next decades.'' (2023-03-06, 86 retweets)\\[2mm]
``These costs have not been caused by \#climate, but by \#green politics. And what will change the climate if Germany conducts \#climate protection alone, while distributing its entire industry to China and the USA? « Ricarda\_Lang: In recent years alone in Germany the climate crisis has caused costs of 145 billion. In the next 20 years it could be up to 900 billion. Nothing is more expensive than not protecting the climate! — »'' (2023-03-07, 54 retweets)
\end{displayquote}

The left-leaning cluster refutes this and emphasizes \node{Germany}'s role in \node{climate protection}:

\begin{displayquote}
  ``China and the USA – these are THE top CO2 producers. Then it doesn’t matter if Germany is doing climate protection and other countries, right? So that’s not right. One thread.'' (2021-08-11, 457 retweets)\\[2mm]
``That's the question at \#btw21, that's the task of our time: "Can we make Germany climate neutral in the next 20 years?" says @ABaerbock in the \#Schlussrunde. We say: yes. We have a plan. We will present a climate protection program for it immediately'' (2021-09-23, 68 retweets)
\end{displayquote}

Another conflict line is activism, and the actions of the \node{Last Generation} in particular. On the right, there are claims that the \node{Last Generation} activists are endangering \node{democracy}:

\begin{displayquote}
``Don't even understand why everyone is so surprised, because the \#Last Generation seems to hate the \#democracy - that's what they say in talk shows, @Luisamneubauer also say. Nobody wanted to see it until now.'' (2023-03-05, 247 retweets)
\end{displayquote}

On the left, the \node{Last Generation} is considered non-\node{violent} and non-criminal:

\begin{displayquote}
``D \#Last Generation calls for consistent climate protection measures to comply with Paris climate targets without using violence, their actions are neither criminal nor terrorist, when will the government act \#Tempolimit immediate stopping of new highway projects \#FDPbelow5Procent « rbb24: The Berlin Public Prosecutor's Office does not currently see any \#initial suspicion of the formation of a criminal association with the "Last Generation". According to a spokesman, the actions are a "lasting insult", but not "like terrorism". — »'' (2023-05-18, 276 retweets)
\end{displayquote}

Continuing to look at conflicting edges, we can examine the relationship between the \node{Greens} and \node{climate change}. We see that the Greens are criticized here by both camps: by the left for not doing enough:

\begin{displayquote}
``I believe two sentences to be true at the same time: 1. The Greens are the only party with a realistic option of power that is committed to climate protection and must therefore be strengthened. 2. The Greens need pressure to put climate protection at the top of the agenda.'' (2023-03-29, 13 retweets)\\[2mm]
``The Greens have simply sacrificed climate protection completely for the FDP. I could freak out, they are such losers... all they needed to do was to insist on existing laws, they can't even do that.... \#Coalition Committee'' (2023-03-28, 11 retweets)
\end{displayquote}

On the right, the \node{Greens} are criticized more vocally of hypocrisy with respect to climate \node{protection}:

\begin{displayquote}
  ``888 The Greens don't care about environmental protection at all! Habeck watches happily as his LNG terminal destroys our Wadden Sea with 480 million litres of chlorinated water every day. The Greens are forcing us, out of pure ideology, to use immature techniques that affect our nature...'' (2023-07-09, 661 retweets)\\[2mm]
``Let us turn to the facts: CO2 emissions are falling in the transport sector, while they are rising again in the energy sector. This also has to do with the blockade of nuclear power. The Greens should finally prefer climate protection to ideology!'' (2023-03-04, 59 retweets)  
\end{displayquote}

Turning to the actors that are only present in one camp, we observe again that on the left, \node{Armin Laschet} is criticized for not taking climate \node{protection} seriously:

\begin{displayquote}
``Armin Laschet demands more climate protection at noon and rejects it exactly four hours later. That is such an opportunistic, stupid rumour-mongering - how could the CDU come up with the idea that someone like that could be a chancellor?'' (2021-07-15, 2855 retweets)  
\end{displayquote}

Furthermore, the \node{FDP} is criticized for not caring about climate \node{protection} and instrumentalizing ``freedom'' when they see it fit:

\begin{displayquote}
``Transport Minister @Wissing lies at \#Maischberger and through a question by "question the state" it turns out that the @FDP minister only met with his Climate Council more than a year after taking office. \#Climate protection is ignored by the FDP. /PM'' (2023-03-24, 769 retweets)\\[2mm]
``At every opportunity, the FDP calls for freedom! Technological freedom should even be incorporated into the law, no matter what it looks like and what it means. But when it comes to letting the cities decide for themselves whether to introduce a speed limit of 30? That’s where Wissing disagrees. Unbelievable.'' (2023-04-22, 126 retweets)  
\end{displayquote}

We also observe that the company responsible for the demolition of Lützerath, \node{RWE}, is not mentioned on the right. The left, on the other hand, is quite vocal about the danger it poses for the \node{climate}:

\begin{displayquote}
  ``The demonstration is on! Thousands of people don't take the RWE climate crime and storm the mine to stop it at last. \#LuetzerathStays \#ClimateDisaster \#PowerToThePeople'' (2023-01-14, 300 retweets)\\[2mm]
``RWE is responsible for ~25\% of CO2 emissions. The government wants to allow this until 2038, even though from a market perspective coal-fired power is already in "death zone" \& renewable energies are MORE ECONOMIC.'' (2020-05-17, 66 retweets)
\end{displayquote}

Furthermore, the left heavily criticizes the \node{state} for lashing out against \node{climate activists}:

\begin{displayquote}
``You can criticize the \#Last Generation for choosing its means, but if the state put all the energy it puts into the fight against climate activists into the fight against the climate catastrophe, we would have no problem.'' (2023-05-24, 2130 retweets)  
\end{displayquote}

We also observe that the left provides a supportive relationship from \node{climate protection} to fundamental \node{rights} and \node{freedom}:

\begin{displayquote}
``The greatest and most appalling irony of our time is that so many people believe that climate protection would restrict us, take away something, end freedom, cost a lot of money and ruin the future, while in all respects the opposite is the case.'' (2023-03-25, 517 retweets)\\[2mm]
``Climate protection does not endanger our freedom, it secures it for the future. \#Bundesverfassungsgericht'' (2021-04-29, 102 retweets)\\[2mm]
  ``I understand this action with color: The \#LastGeneration wants to draw attention to the fact that too little \#ClimateProtection jeopardizes our \#FundamentalRights. It calls on the government to adhere to the laws in force. \#KSG Why should the LG hate our GG [GrundGesetz: constitution]? « AufstandLastGen: ++ Berlin: Monument of Fundamental Rights engraved in “Erdöl” ++ The artwork near the Bundestag building shows the articles of the Basic Law. Today we have shown how the government deals with these. Burn gas or protect fundamental rights? In 2023, only one of both is possible. — »'' (2023-03-05, 108 retweets)
\end{displayquote}

Looking at actors and relationships only mentioned on the right side, we observe that the \node{media} is blamed for over-emphasizing \node{climate change} and for not warning the population early enough of the flood disasters:

\begin{displayquote}
  ``While the media repeatedly seeks to blame the "climate change" for the current forest fires in Canada, other reports show that it is primarily arsonists who are behind them.'' (2023-06-11, 213 retweets)\\[2mm]
``The \#flood disaster shows above all the inability of politics and the failure of the public media to warn of the disaster. A country that deals with gender and the rescue of the world is incapable of solving real problems.'' (2021-07-18, 206 retweets)
\end{displayquote}

The \node{authorities} too are blamed for not warning about the \node{floods} early enough:

\begin{displayquote}
``Authorities, politicians and public broadcasters apparently did not want to warn people of the flood.'' (2021-07-19, 86 retweets)
\end{displayquote}

Finally, we find one central issue on the right, which is the relationship between human-induced \node{CO\textsubscript{2}} emissions and \node{climate change}:

\begin{displayquote}
  ``There is no “man-made” climate change. But there is a man-made \#ClimateScam that benefits its “inventors” alone.'' (2023-03-07, 205 retweets)\\[2mm]
  ``The Earth’s climate has always changed at all times. “Climate change” is a banality. That the climate is man-made for an imminent catastrophe is a myth that was invented to enforce a dictatorship. \#Climate change \#Climate catastrophe'' (2023-03-29, 75 retweets)\\[2mm]
  ``I have learned in my physics studies that any model that wants to describe a process in nature has to be tested experimentally. I have been looking for months for experiments that confirm the theory that CO2 warms the climate. I find nothing.'' (2023-05-20, 475 retweets)
\end{displayquote}

\end{document}